\documentclass[10pt,twocolumn,letterpaper]{article}

\usepackage{iccv}              % To produce the CAMERA-READY version
\usepackage[accsupp]{axessibility}
\definecolor{iccvblue}{rgb}{0.21,0.49,0.74}
\usepackage[pagebackref,breaklinks,colorlinks,allcolors=iccvblue]{hyperref}
\usepackage{times}
\usepackage{epsfig}
\usepackage{graphicx}
\usepackage{amsmath}
\usepackage{amssymb}
\usepackage{booktabs}
\usepackage{setspace}
\usepackage{multirow}
\usepackage{tikz} % To generate the plot from csv
\usepackage{pgfplots}
\pgfplotsset{compat=1.5}

\usepackage{enumitem}
\setlist[itemize]{noitemsep,leftmargin=*,topsep=0em}
\setlist[enumerate]{noitemsep,leftmargin=*,topsep=0em}

\usepackage{microtype}

\usepackage{inconsolata}

\usepackage{url}
\usepackage{amsmath}
\usepackage{adjustbox}
\usepackage{booktabs}
\usepackage{multirow}
\usepackage{makecell}
\usepackage{graphicx}
\usepackage{tikz}
\usepackage{pgfplots}
\usepackage{enumitem}
\usepackage{amssymb}
\usepackage{pifont}
\usepackage{xcolor}
\usepackage[most]{tcolorbox}
\usepackage{tcolorbox}
\usepackage{colortbl}
\usepackage{soul}

\usepackage{siunitx}
\usetikzlibrary{shadows}
\usetikzlibrary{plotmarks}
\usetikzlibrary{patterns}
\usetikzlibrary{pgfplots.groupplots}
\usetikzlibrary{arrows}
\usetikzlibrary{decorations}
\usetikzlibrary{decorations.pathreplacing}
\usetikzlibrary{backgrounds}
\usetikzlibrary{fit}
\usetikzlibrary{positioning}
\usetikzlibrary{calc}
\usetikzlibrary{shapes.multipart}

\usepackage{caption}

\definecolor{ugreen}{cmyk}{1,0,1,0.498}
\definecolor{lyyblue}{cmyk}{0.8278,0.3333,0,0.2941}
\definecolor{lyygreen}{cmyk}{0.6813,0,0.725,0.3725}
\definecolor{lyyred}{cmyk}{0,0.8855,0.8767,0.1098}
\definecolor{dblue}{cmyk}{1,0.5487,0,0.5569}

\definecolor{myred}{HTML}{E33222}

\definecolor{my_green}{RGB}{51,102,0}
\definecolor{my_red}{RGB}{204, 0, 0}

\definecolor{mypurple}{RGB}{147,112,219}
\definecolor{gr}{RGB}{0, 146, 0}

\newcommand{\benchmark}{EgoMask}
\newcommand{\trainingdataset}{EgoMask-Train}
\newcommand{\TRecall}{T_{recall}}
\newcommand{\IouAll}{IoU_{all}}
\newcommand{\IouGold}{IoU_{gold}}
\newcommand{\IouGoldPred}{IoU_{gold\_pred}}

\newcommand\blfootnote[1]{%
  \begingroup
  \renewcommand\thefootnote{}\footnote{#1}%
  \addtocounter{footnote}{-1}%
  \endgroup
}

\definecolor{navyblue}{HTML}{0071BC}
\definecolor{hotpink}{HTML}{FF0080}

\definecolor{oai-white}{HTML}{FFFFFF}
\definecolor{oai-black}{HTML}{000000}
\definecolor{oai-red}{HTML}{FF4500}
\definecolor{oai-green}{HTML}{51DA4C}
\definecolor{oai-blue}{HTML}{0000FF}
\definecolor{oai-yellow}{HTML}{FFF639}
\definecolor{oai-magenta}{HTML}{FF45FF}
\definecolor{oai-cyan}{HTML}{00FFFF}
\definecolor{oai-orange}{HTML}{FE7600}
\definecolor{oai-violet}{HTML}{8A2BE2}
\definecolor{oai-brown}{HTML}{A0522D}
\definecolor{oai-green-050}{HTML}{F4FFF4}
\definecolor{oai-green-100}{HTML}{E9FFE8}
\definecolor{oai-green-200}{HTML}{D9FFD8}
\definecolor{oai-green-300}{HTML}{C9FFC7}
\definecolor{oai-green-400}{HTML}{A6FFA3}
\definecolor{oai-green-500}{HTML}{7CF178}
\definecolor{oai-green-600}{HTML}{51DA4C}
\definecolor{oai-green-700}{HTML}{3FA93B}
\definecolor{oai-green-800}{HTML}{2D712A}
\definecolor{oai-green-900}{HTML}{193718}
\definecolor{oai-gray-000}{HTML}{FFFFFF}
\definecolor{oai-gray-100}{HTML}{FAFAFA}
\definecolor{oai-gray-200}{HTML}{F5F5F5}
\definecolor{oai-gray-300}{HTML}{E5E5E5}
\definecolor{oai-gray-400}{HTML}{FFB7A4}
\definecolor{oai-gray-500}{HTML}{CDCDCD}
\definecolor{oai-gray-600}{HTML}{A8A8A8}
\definecolor{oai-gray-700}{HTML}{747474}
\definecolor{oai-gray-800}{HTML}{393939}
\definecolor{oai-gray-900}{HTML}{000000}
\definecolor{visual}{HTML}{A50E0E}       
\definecolor{linguistic}{HTML}{174EA6}   
\definecolor{relational}{HTML}{E37400}   
\definecolor{egocentric}{HTML}{0D652D}  
\colorlet{mapcolor}{ForestGreen}

\newcommand{\infobox}[1]{
    \vspace{-0.18cm}
    \begin{tcolorbox}[
        colback=white!90!gray,     
        colframe=teal!60!black,   
        arc=5pt,                   
        boxsep=5pt,                 
        left=5pt,                  
        right=10pt,                 
        top=2pt,                   
        bottom=3pt,                
        boxrule=0.8pt,              
        drop shadow=gray!50!white, 
        enhanced jigsaw             
    ]
    \vspace{-0.1cm}
         \textit{#1}
    \vspace{-0.2cm}
    \end{tcolorbox}
    \vspace{-0.15cm}
}

\newcommand{\cmark}{\textcolor{my_green}{\ding{51}}} % ✔
\newcommand{\xmark}{\textcolor{orange}{\ding{55}}} % ✘

\def\paperID{5312} % *** Enter the Paper ID here
\def\confName{ICCV}
\def\confYear{2025}

\title{Fine-grained Spatiotemporal Grounding on Egocentric Videos}

\author{
Shuo Liang,
Yiwu Zhong,
Zi-Yuan Hu,
Yeyao Tao,
Liwei Wang$^{\dagger}$ \\
The Chinese University of Hong Kong %\quad
}

\begin{document}
\maketitle

\blfootnote{ $^\dagger$ Corresponding author}

\begin{abstract}

Spatiotemporal video grounding aims to localize target entities in videos based on textual queries. While existing research has made significant progress in exocentric videos, the egocentric setting remains relatively underexplored, despite its growing importance in applications such as augmented reality and robotics.
In this work, we conduct a systematic analysis of the discrepancies between egocentric and exocentric videos, revealing key challenges such as shorter object durations, sparser trajectories, smaller object sizes, and larger positional shifts. 
To address these challenges, we introduce \benchmark{}, the first pixel-level benchmark for fine-grained spatiotemporal grounding in egocentric videos. It is constructed by our proposed automatic annotation pipeline, which annotates referring expressions and object masks across short-, medium-, and long-term videos. 
Additionally, we create \trainingdataset{}, a large-scale training dataset to facilitate model development. 
Experiments demonstrate that the state-of-the-art spatiotemporal grounding models perform poorly on our benchmark \benchmark{}, but fine-tuning on \trainingdataset{} yields significant improvements, while preserving performance on exocentric datasets. Our work thus provides essential resources and insights for advancing egocentric video understanding. Our code is available at \href{https://github.com/LaVi-Lab/EgoMask}{\textcolor{magenta}{https://github.com/LaVi-Lab/EgoMask}}.
% \url{https://github.com/LaVi-Lab/EgoMask}.
% \textit{\url{https://github.com/LaVi-Lab/EgoMask}}.

\end{abstract}    
\section{Introduction}
\label{sec:intro}

\begin{figure}[t]
\begin{center}
\includegraphics[width=1.0\linewidth]{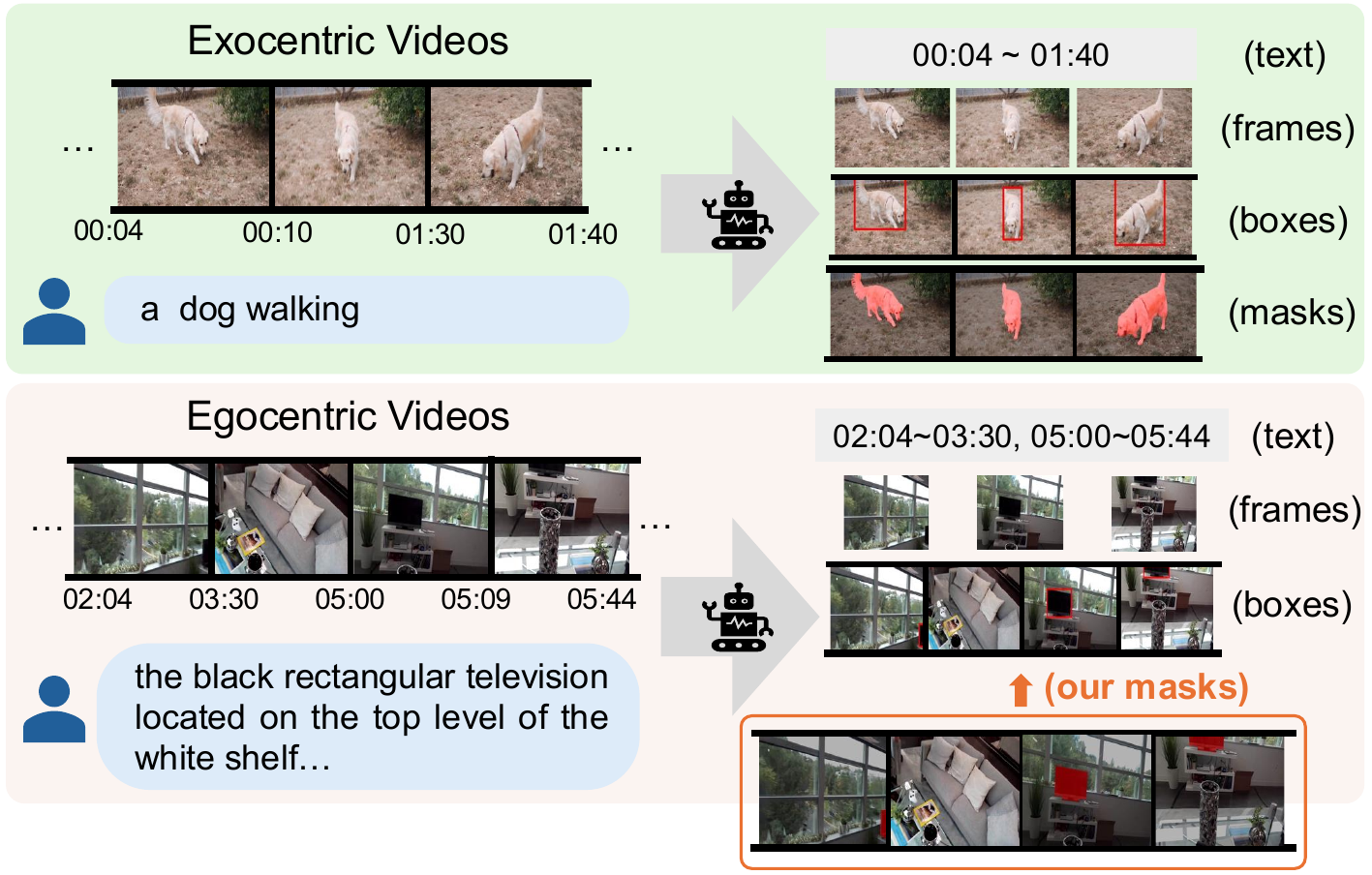}
\end{center}
\vspace{-20pt}
   \caption{Comparison of video grounding tasks. We propose the first pixel-level benchmark for fine-grained spatiotemporal grounding in egocentric videos.
   } 
   \vspace{-10pt}
\label{fig:teaser_figure}
\end{figure}

Video grounding has been extensively studied in recent years~\cite{DBLP:conf/eccv/SeoLH20, 
DBLP:conf/cvpr/ZhangZZWLG20,
DBLP:journals/corr/abs-2107-09609,
DBLP:conf/cvpr/GraumanWBCFGH0L22,
DBLP:conf/nips/JinLYM22,
DBLP:journals/tcsv/TangLLLJJYX22,
DBLP:conf/cvpr/Gu00L024}. 
It requires models to generate a temporal tube for target entities based on a given language query.
As illustrated in Figure~\ref{fig:teaser_figure}, this tube can be represented as a sequence of consecutive frames~\cite{DBLP:journals/corr/abs-2107-09609,DBLP:conf/cvpr/GraumanWBCFGH0L22} (temporal grounding), or a set of bounding boxes and masks that localize entities within these frames~\cite{DBLP:conf/eccv/SeoLH20,DBLP:conf/cvpr/Gu00L024} (spatiotemporal grounding). The mask output further requires a pixel-level localization on the fine-grained object geometry.
It is worth noting that existing works~\cite{DBLP:journals/corr/abs-2501-04001, DBLP:conf/nips/Bai0MWGClZS24} predominantly focus on spatiotemporal grounding in exocentric videos~\cite{DBLP:conf/iccv/Ding0H0L23,DBLP:conf/eccv/SeoLH20}, while egocentric video grounding remains relatively underexplored.
This is despite the growing interest in egocentric video understanding~\cite{DBLP:conf/cvpr/DiX24,DBLP:journals/corr/abs-2410-07177,DBLP:conf/nips/ChandrasegaranG24}, driven by its potential applications in augmented reality (AR) glasses, household robotics, and other real-world scenarios.

Meanwhile, existing studies have observed a gap between egocentric and exocentric videos~\cite{DBLP:conf/cvpr/GraumanWBCFGH0L22, DBLP:conf/nips/TangLGF023, DBLP:conf/iccv/KuritaKO23}. For instance, egocentric videos are characterized by rapid camera movements, causing objects to frequently enter and exit the field of view, while also undergoing rapid appearance changes.
However, a systematic and statistical analysis of these differences remains lacking.
Therefore, we conducted a study that quantitatively analyzes the discrepancies between egocentric and exocentric videos. Our findings reveal several key differences: entities in egocentric videos exhibit \textbf{shorter} total durations, \textbf{sparser} continuous trajectories, \textbf{smaller} object sizes, and \textbf{larger} positional shifts compared to their counterparts in exocentric videos. This study raises a key question: {\textit {Can existing models perform well on fine-grained spatiotemporal grounding in egocentric videos?}}

To answer the question, however, there lacks a benchmark specifically designed for fine-grained spatiotemporal grounding in egocentric videos. The most relevant datasets are EgoTracks~\cite{DBLP:conf/nips/TangLGF023} and RefEgo~\cite{DBLP:conf/iccv/KuritaKO23}. 
EgoTracks focuses on long-term object tracking in egocentric videos but only provides object category labels and bounding box annotations. RefEgo, on the other hand, evaluates referring expression comprehension in egocentric videos and includes text queries, yet it also provides only bounding box annotations.
Importantly, RefEgo consists solely of short video segments (under 1 minute), which do not align with the typical visual inputs in egocentric AI applications, where models process long egocentric video streams.

In this work, we propose an automatic annotation pipeline to construct a benchmark that provides pixel-level annotations and enables the evaluation of spatiotemporal grounding in egocentric videos across various durations.
To generate mask annotations, we leverage the pre-trained segmentation model SAM2~\cite{DBLP:journals/corr/abs-2408-00714}, using the bounding boxes provided by the EgoTracks dataset. 
To generate object expressions as language queries, we employ the vision-language model GPT-4o~\cite{gpt4o} through two strategies to ensure data diversity: (1) prompting GPT-4o to directly generate both short and long expressions, and (2) first prompting GPT-4o to produce metadata about the target object, such as visual attributes and world knowledge, which
is then combined to construct referring expressions.
Finally, both mask and expression annotations are refined and verified by human annotators.
As a result, we introduce the first egocentric, fine-grained spatiotemporal grounding benchmark, \benchmark{}, built upon a subset of EgoTracks and RefEgo. The benchmark comprises 700 queries across 315 videos, spanning \textbf{short-term} (under 1 minute), \textbf{medium-term} (1 to 3 minutes), and \textbf{long-term} (over 3 minutes) durations to enable a comprehensive evaluation.
Moreover, to support model training, we create \textbf{\textit{\trainingdataset{}}}, a large-scale training set created by our automatic pipeline. It includes 2,624 videos with mask annotations for 9,592 objects and a total of 47,968 referring expressions.

With our newly proposed benchmark, we systematically evaluate the state-of-the-art (SOTA) spatiotemporal grounding models on egocentric videos.
Experiment results reveal that existing SOTA models~\cite{DBLP:journals/corr/abs-2501-04001,DBLP:conf/nips/Bai0MWGClZS24} perform significantly worse on our benchmark compared to their performance on existing exocentric benchmarks. 
Going further, we fine-tune pre-trained SOTA models on our created training dataset, \textit{\trainingdataset{}}, leading to large performance improvements (\eg, an average relative increase of 41.30\%). These results highlight the effectiveness of our dataset as a valuable resource for advancing egocentric spatiotemporal grounding.
Notably, we observe that models fine-tuned on our egocentric dataset still retain their performance on existing exocentric benchmarks. This suggests that our dataset is complementary to existing exocentric datasets and can serve as a unique and orthogonal data source for training future video foundation models.
In addition to end-to-end trained SOTA models, we also evaluate a baseline method, Grounded-SAM2~\cite{ren2024grounded}, a pipeline framework that first runs GrondingDino~\cite{liu2024grounding} to localize objects based on text queries, followed by SAM2~\cite{DBLP:journals/corr/abs-2408-00714} for object tracking. Even though it uses the same tracking model as our annotation pipeline, this method also fails to perform well on egocentric videos, which further verifies the difficulty of our proposed benchmark and the challenges of fine-grained spatiotemporal grounding on egocentric videos.

Overall, our contributions are as follows:

\begin{itemize}
    \item We explore spatiotemporal video grounding for egocentric videos and develop an automatic data annotation pipeline, resulting in the first pixel-level benchmark \textit{\benchmark{}} and a large-scale training dataset \textit{\trainingdataset{}}.
    \item We conduct in-depth analysis to quantitatively and systematically measure the gap between exocentric and egocentric videos, providing insights for future modeling. 
    \item Extensive experiments reveal that existing spatiotemporal grounding models fail to perform effectively on egocentric videos, and our collected training data can remarkably improve existing models.
\end{itemize}
\section{Related work}
\label{sec:related_work}

\noindent \textbf{Referring Video Segmentation Dataset.}
Referring video object segmentation (RVOS)~\cite{gavrilyuk2018actor,mttr,wu2022referformer,wu2023onlinerefer,li2023robust,miao2023spectrum}, a subtask of spatiotemporal video grounding. 
It aims to segment the target objects in a video based on a given natural language expression. Compared to standard video object segmentation~\cite{DBLP:journals/tist/YaoLXZZ20}, RVOS is more challenging as it requires models to effectively integrate both visual and linguistic information.
The RVOS task was first introduced by~\cite{gavrilyuk2018actor} along with the A2D-Sentences and J-HMDB Sentences datasets, which primarily focus on actor segmentation. Subsequent works~\cite{DBLP:conf/accv/KhorevaRS18,DBLP:conf/eccv/SeoLH20,DBLP:conf/iccv/Ding0H0L23,DBLP:conf/nips/Bai0MWGClZS24} have expanded the scope of RVOS by scaling up datasets, increasing task complexity, or enhancing task diversity.
Specifically, Ref-DAVIS~\cite{DBLP:conf/accv/KhorevaRS18} extends the DAVIS datasets~\cite{DBLP:conf/cvpr/PerazziPMGGS16,DBLP:journals/corr/Pont-TusetPCASG17} by replacing segmentation masks with language descriptions, introducing a multi-object segmentation setting. Given its relatively small scale, Refer-YouTube-VOS~\cite{DBLP:conf/eccv/SeoLH20} was built on YouTube-VOS~\cite{DBLP:journals/corr/abs-1809-03327}, featuring diverse object categories and longer videos. MeViS~\cite{DBLP:conf/iccv/Ding0H0L23} incorporates more complex scenes with a higher density of objects, with expressions focusing on object motion.
ReasonVOS~\cite{DBLP:conf/nips/Bai0MWGClZS24} emphasizes complex reasoning over language queries and temporal object tracking with explicit motion understanding.
However, these datasets predominantly feature exocentric videos which are generally shorter and exhibit less rapid camera movement. In contrast, we focus on egocentric videos which present additional challenges due to their longer durations, frequent and large camera movements, and increased object density. 

\smallskip
\noindent \textbf{Egocentric Dataset.} Egocentric video understanding~\cite{DBLP:conf/cvpr/GraumanWBCFGH0L22,egovlp,egovlpv2,egovideo} has recently emerged as a pivotal research area in Embodied AI~\cite{DBLP:journals/ijcv/PlizzariGFBRFDT24,EmbodiedGPT}, presenting challenges distinct from those in exocentric video analysis. Unlike conventional exocentric datasets~\cite{jang2017tgif, xu2017video, xiao2021next, yu2019activitynet}, egocentric videos are captured from a first-person perspective using wearable devices, resulting in long-form, dynamic footage characterized by frequent camera motion.
To advance egocentric video understanding, numerous datasets have been introduced~\cite{DBLP:conf/cvpr/GraumanWBCFGH0L22,egoschema,QAEGO4D,DBLP:conf/cvpr/DiX24,ego4dgoalstep,DBLP:conf/nips/TangLGF023,DBLP:conf/iccv/KuritaKO23,VISOR2022,perrett2025hdepic}. 
Ego4D~\cite{DBLP:conf/cvpr/GraumanWBCFGH0L22} laid the foundation by providing a large-scale, comprehensive egocentric dataset. Building upon Ego4D, several datasets have been developed to extend the scope of egocentric video understanding. For instance, EgoSchema~\cite{egoschema} focuses on long-form egocentric video question-answering (VideoQA), while~\cite{QAEGO4D,DBLP:conf/cvpr/DiX24} addresses grounding in VideoQA. EgoTracks~\cite{DBLP:conf/nips/TangLGF023} is designed for long-term object tracking, and RefEgo~\cite{DBLP:conf/iccv/KuritaKO23} evaluates referring expression comprehension.
However, these datasets do not support pixel-level spatiotemporal grounding due to the lack of mask annotations. 
% EPIC-Visor [revised]
EPIC-Visor~\cite{VISOR2022} provides pixel-level masks. However, it has only 158 videos that are in kitchens, and its clips with consistent mask annotations are 12 seconds on average, failing to support long video grounding.
% HD-EPIC [newly-added]
The subsequent work HD-EPIC~\cite{perrett2025hdepic} provides pixel-level segmentations for longer object movement tracks, which also only focuses on kitchen activities and lacks referring expressions towards the labeled objects.
In contrast, our \benchmark{} provides diverse textual queries alongside mask annotations, covering video lengths ranging from seconds to minutes. This facilitates more fine-grained and comprehensive spatiotemporal grounding.

% TODO [need revise]
\smallskip
\noindent \textbf{Multimodal Large Language Model.}
Multimodal large language models (MLLMs)~\cite{llava15,qwen2vl,gpt4o} have made significant progress in visual-language tasks~\cite{hudson2019gqa,yue2023mmmu,yu2023mm}.
Recently, many studies explore their capabilities in pixel-level understanding~\cite{zhang2024llava,DBLP:conf/cvpr/LaiTCLY0J24,DBLP:conf/nips/Bai0MWGClZS24,DBLP:journals/corr/abs-2501-04001}. 
A common practice is using special tokens to enable MLLMs with grounding ability, such as the Segment Anything Model~(SAM)~\cite{DBLP:conf/iccv/KirillovMRMRGXW23, DBLP:journals/corr/abs-2408-00714}. 
LISA~\cite{DBLP:conf/cvpr/LaiTCLY0J24} and VideoLISA~\cite{DBLP:conf/nips/Bai0MWGClZS24} introduce a special ``[SEG]'' token to connect the MLLM with SAM and perform independent image segmentation, while Sa2VA~\cite{DBLP:journals/corr/abs-2501-04001} integrate SAM2~\cite{DBLP:journals/corr/abs-2408-00714} into MLLMs, allowing the utilization of other frames to perform video segmentation.
%
% A common practice is using special tokens to enable MLLMs with grounding ability, such as the Segment Anything Model~(SAM)~\cite{DBLP:conf/iccv/KirillovMRMRGXW23, DBLP:journals/corr/abs-2408-00714}. 
%
% LISA~\cite{DBLP:conf/cvpr/LaiTCLY0J24} uses a special ``[SEG]'' token to connect the MLLM with SAM, introducing the embedding-as-mask paradigm to unlock the segmentation capability of MLLMs, a concept that has inspired many subsequent works.
% %
% However, LISA only supports the segmentation of a single image. 
% Based on its model architecture,
% VideoLISA~\cite{DBLP:conf/nips/Bai0MWGClZS24} takes a step further by utilizing a single special token to perform video object segmentation.
% It can be viewed as performing independent image segmentation on the video frames.
% %
% However, such method may suffer from inconsistent segmentation due to neglecting the information from other frames.
% %
% To overcome this issue,
% Sa2VA~\cite{DBLP:journals/corr/abs-2501-04001} integrate SAM2~\cite{DBLP:journals/corr/abs-2408-00714}, a foundation model towards solving segmentation in images and videos, into MLLMs.
% It uses the ``[SEG]'' token to generate the masks of key frames and uses the memory module from SAM2 to predict on the remaining frames.
%
However, these models primarily focus on exocentric videos. It is still unclear whether they can perform well on egocentric videos.
In this work, we create a benchmark to evaluate them on egocentric spatiotemporal grounding and a training dataset to enhance their performance.
The fine-tuned models achieve large improvements on egocentric videos, without performance loss on exocentric videos. 

\section{Method}
Our work fills a critical gap in pixel-level spatiotemporal grounding for egocentric videos, better supporting egocentric applications in augmented reality and robotics. Specifically, we design an automatic pipeline that utilizes existing segmentation models to generate pixel-level object masks and leverages visual-language models to produce object expressions as text queries. This pipeline enables the construction of both a training dataset and an evaluation benchmark, laying the foundation for advancing egocentric video understanding.

% In order to construct a pixel-level spatiotemporal grounding dataset for better supporting egocentric applications in the AR and robotics domain, we build an automatic pipeline to employ existing segmentation tools to collect pixel-level annotations and use visual-language models to generate object expressions as text queries. Based on the above pipeline, we collect a training dataset and a benchmark for supporting fine-grained spatiotemporal grounding on egocentric videos.

\begin{figure}[t]
\begin{center}
\includegraphics[width=0.99\linewidth]{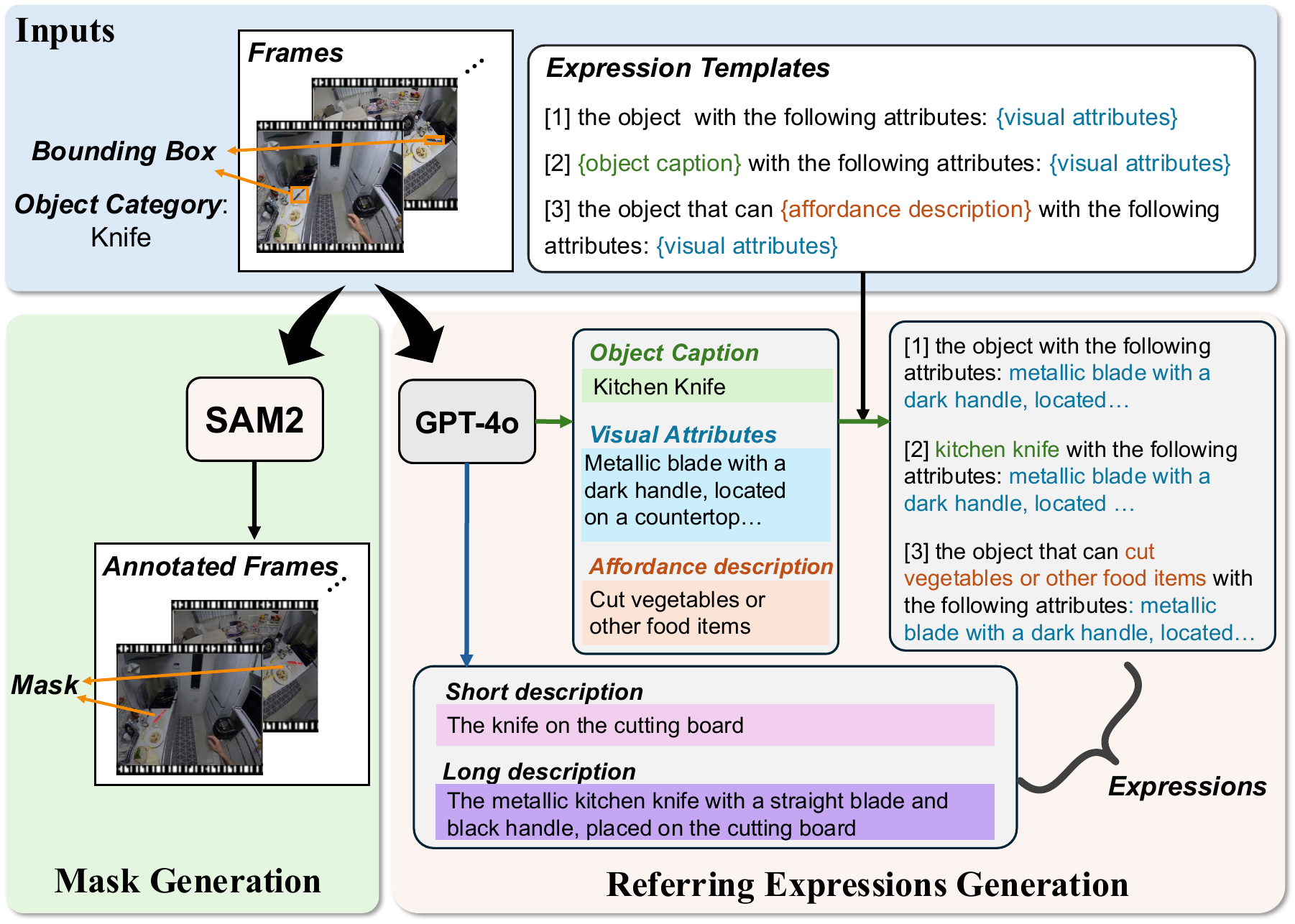}
\end{center}
\vspace{-15pt}
   \caption{Automatic Annotation Pipeline. The inputs include video frames, the annotated bounding boxes, and the object category from the Egotracks dataset. The annotation process contains two parts: (1) \textbf{Mask Generation} (Bottom Left): utilizing SAM2 as annotator to generate mask throughout the input frames; (2) \textbf{Referring Expression Generation} (Bottom Right): prompting GPT-4o to directly generate referring expressions (blue arrow) or first generate metadata about the labeled object and then adopt pre-defined templates to generate expressions (green arrow).
   } 
   % \caption{Automatic annotation pipeline, which takes the frames annotated with bounding box and the corresponding object category from Egotracks dataset as the input and contains two parts: (1) \textbf{Mask Generation} (Bottom Left): utilizing SAM2 as annotator to generate mask throughout the input frames; (2) \textbf{Referring Expression Generation} (Bottom Right): prompting GPT-4o to directly generate referring expressions (blue arrow) or firstly generate metadata about the labeled objects and then adopt pre-defined templates to form expressions (green arrow).
   % } 
   \vspace{-5pt}
\label{fig:anno_pipeline}
\end{figure}

\subsection{Automatic Annotation Pipeline}

Existing egocentric datasets are not directly suitable for pixel-level spatiotemporal grounding tasks. For instance, EgoTracks~\cite{DBLP:conf/nips/TangLGF023} provides long-term object tracking with only bounding box annotations and lacks referring expressions for the annotated objects.
Given the high cost of manually annotating object masks, we build upon the densely annotated bounding boxes in EgoTracks and introduce additional annotations to address the absence of pixel-level masks and language queries. Specifically, we propose an automatic annotation pipeline that: (1) generates object masks using existing segmentation tools guided by the provided bounding boxes, and (2) produces referring expressions (\ie, language queries) for the target objects using vision-language models, as illustrated in Figure~\ref{fig:anno_pipeline}.

% Existing egocentric datasets can not directly support pixel-level spatiotemporal grounding tasks. For example, the Egotracks~\cite{DBLP:conf/nips/TangLGF023} provides long-term object-tracking information with only bounding box annotations. Also, there are no referring expressions towards the annotated objects. 
% Taking the human efforts to annotate masks into consideration, we resort to using the dense human-annotated bounding box information from Egotracks and conduct advanced annotations to compensate for the lack of object masks and referring expressions.
% Thus, we propose an automatic annotation pipeline to: (1) generate mask annotation via existing segmentation tools guided by the given bounding boxes; (2) generate referring expressions (language queries) of the target object by using vision-language models, as illustrated in Figure~\ref{fig:anno_pipeline}.

\smallskip \noindent \textbf{Pixel-level mask generation.}  In order to reduce the annotation time and avoid segmentation errors caused by object absence, we only select the clip segments that contain the target object to annotate. 
For each clip segment, the bounding box in the first frame is selected as the box prompt input of SAM2~\cite{DBLP:journals/corr/abs-2408-00714}, a powerful unified model that can segment objects across images and generate their masks throughout the video clip. 
%
% We then post-process the generated masks by only keeping the areas that overlap with box annotations to avoid unexpected segmentation results further.
%
We then post-process the generated masks by only keeping the areas that overlap with box annotations.
In this way, we can minimize hallucination errors and ensure that the generated masks are within or near the area of the target objects. %, which guarantees the generation quality.

\smallskip
\noindent \textbf{Referring expression generation.} In grounding tasks, the referring expression should correspond to the unique target object without any ambiguity, which requires the annotation tool to fully understand the distinction of the target object from the surroundings. 
Furthermore, for better supporting egocentric applications in complex situations, the referring expressions should cover diverse aspects of the objects~\cite{zhong-2024-beyond}, including simple object captions, various visual attributes, and world knowledge about the objects.
Thus, to guarantee the \textit{diversity} of referring expressions, we propose two different strategies to generate  referring expressions. 
\begin{itemize}
    \item Prompting GPT-4o to directly generate a short description and a long description. 
    We first select three frames that have the most clear objects (\eg, large bounding boxes) from the video and make sure they come from different trajectories. 
    Then, We draw boxes of the target object on video frames and prompt GPT-4o to generate descriptions.
    \item Prompting GPT-4o to generate metadata first and then produce expressions based on designed templates.
    Inspired by~\cite{zhong-2024-beyond}, the metadata includes object caption, visual attributes (\eg, physical locations, color, shape, dynamics), and world knowledge (\eg, object affordance). 
    Once these metadata are obtained, we utilize the templates shown at the top of Figure~\ref{fig:anno_pipeline} to form expressions.
\end{itemize}
The detailed prompts can be found in the Appendix.

% \usepackage{graphicx}
\begin{table*}[htb]
\centering
\resizebox{1.0\textwidth}{!}{%
\begin{tabular}{c|cc|c|ccc|c|cccc}
\toprule
% \multirow{2}{*}{} 
\multirow{2}{*}{\textit{\textbf{Training dataset}}} &
\textbf{Video} &
\textbf{Total}  &
\textbf{Mask}  &
\multirow{2}{*}{\textbf{\#Traj.}} &
\textbf{Avg. Traj.}  &
\textbf{Disappear}  &
\textbf{Adj. Mask} &
\multirow{2}{*}{\textbf{\#Video}} &
\multirow{2}{*}{\textbf{\#Object}} &
\multirow{2}{*}{\textbf{\# Expr.}} &
\textbf{Avg. Expr.} \\

                                &  \textbf{Length (s)} & \textbf{Duration(\%)} & \textbf{Area(\%)} & & \textbf{Length(\%)}  & \textbf{Ratio(\%)} &  \textbf{IoU(\%)} & & &   & \textbf{Length} \\
\midrule

\multicolumn{12}{c}{\textit{\textbf{Existing Exocentric Training Dataset}}} \\ 
\midrule

Ref-Davis & 71.18  & 94.33 & 5.84  & 1.22  & 87.45 & 5.85   & 64.96 & 60   & 142  & 572   & 6.27  \\
Mevis     & 69.67  & 77.51 & 5.34  & 1.43  & 68.52 & 20.29  & 54.83 & 1,662 & 6,719 & 23,051 & 7.05  \\
Ref-YT-VOS & 26.53  & 93.57 & 10.23 & 1.11  & 89.65 & 5.97   & 61.62 & 3,471 & 6,459 & 12,913 & 9.63 \\
\midrule
\multicolumn{12}{c}{\textit{\textbf{Our Egocentric Training Dataset}}} \\ 
\midrule

\textbf{\trainingdataset{}}  & 369.94 & 21.56 & 1.20  & 17.00 & 1.33  & 655.82 & 14.96 & 2,624 & 9,592 & 47,968 & 21.67 \\
\bottomrule

\end{tabular}%
}
\caption{The statistics of our proposed training dataset and the comparison with existing egocentric training datasets. 
The ``\textbf{Total Duration}~(\%)'' means the percent of the total appearance of the objects. 
The ``Mask Area~(\%)'' means the average area of the annotated mask over the frame size, which can reveal the \textbf{object size}.
The ``\# Traj.'' means the number of object's continuous trajectories throughout the video, where the trajectory is defined as one consecutive appearance in the video.
The  ``Avg. Traj. Length (\%)'' means the average of each trajectory duration over the whole video and
the ``Disappear. Ratio(\%)'' is formulated as the mean of disappearance duration over trajectory duration.
These two values can reveal \textbf{the sparsity of the continuous trajectory}.
the ``Adj. Mask IoU(\%)'' shows the \textbf{positional shifts} over the adjacent frames by calculating the IoU value of the masks of the target object.
The notion ``Expr.'' refers to the expressions.
}
\label{table:training-dataset}
\end{table*}

% \usepackage{multirow}
% \usepackage{graphicx}
\begin{table*}[htb]
\centering
\resizebox{1.0\textwidth}{!}{%
\begin{tabular}{cc|cc|c|ccc|c|cccc}
\toprule
% \multirow{2}{*}{} 
% \multicolumn{2}{c|}{\blue{\textbf{Benchmark}}} &
\multicolumn{2}{c|}{\multirow{2}{*}{\textit{\textbf{Benchmark}}}} &
\textbf{Video} &
\textbf{Total}  &
\textbf{Mask}  &
\multirow{2}{*}{\textbf{\#Traj.}} &
\textbf{Avg. Traj.}  &
\textbf{Disappear}  &
\textbf{Adj. Mask} &
\multirow{2}{*}{\textbf{\#Video}} &
\multirow{2}{*}{\textbf{\#Object}} &
\multirow{2}{*}{\textbf{\#Expr.}} &
\textbf{Avg. Expr.} \\

                                & &  \textbf{Length (s)} & \textbf{Duration(\%)} & \textbf{Area(\%)} & & \textbf{Length(\%)}  & \textbf{Ratio(\%)}&  \textbf{IoU(\%)} & & &   & \textbf{Length} \\
\midrule
\multicolumn{13}{c}{\textit{\textbf{Existing Exocentric Benchmark}}} \\ 
\midrule
\multicolumn{2}{c|}{Ref-Davis}   & 65.31  & 98.77 & 5.92    & 1.10  & 95.88 & 1.09   & 71.13 & 30  & 61  & 244 & 6.11  \\

\multicolumn{2}{c|}{Mevis}   & 76.55  & 88.14 & 4.92    & 1.36  & 74.77 & 11.44  & 60.82 & 50  & 198 & 793 & 7.70  \\

\multicolumn{2}{c|}{ReasonVOS}    & 98.12  & 89.34 & 1247.75 & 1.34  & 72.03 & 15.73  & 66.61 & 91  & 105 & 458 & 14.09 \\
\midrule

\multicolumn{13}{c}{\textit{\textbf{Our Egocentric Benchmark}}} \\ 
\midrule

\multirow{3}{*}{\textbf{\benchmark{}}} & Short & 12.15  & 80.31 & 1.83    & 1.66  & 57.13 & 21.92  & 8.51  & 200 & 200 & 400 & 13.27 \\
                         & Medium   & 116.30 & 36.69 & 1.87    & 4.62  & 11.10 & 187.84 & 21.15 & 100 & 100 & 200 & 17.34 \\
                         & Long  & 361.32 & 27.48 & 1.86    & 14.60 & 1.81  & 450.29 & 19.53 & 15  & 50  & 100 & 17.34 \\
\bottomrule
\end{tabular}%
}
\vspace{-5pt}
\caption{The statistics of our proposed benchmark and comparison with existing exocentric benchmarks.}
\label{table:benchmark-test}
\vspace{-5pt}
\end{table*}

\subsection{Dataset Curation}
Our automatic pipeline is used to construct both an evaluation set and a training set. The evaluation set undergoes additional verification by human annotators to ensure annotation quality, while the training set is used to train spatiotemporal grounding models.

% Based on our automatic pipeline, we collect training data based on the subset of the Egotracks. We also use this pipeline to generate the noisy test data for further human verification, which largely improves the annotation efficiency.

\smallskip \noindent \textbf{Training dataset curation.} 
Our training set \textbf{\textit{\trainingdataset}} comes from the filtered subset of Egotracks. It contains 2,624 videos and mask annotations of 9,592 objects with 47,968 expressions. 
The detailed statistics are shown in Table~\ref {table:training-dataset}, alongside the comparison with the existing exocentric training dataset.
Note that we annotate the video at 1 FPS, while the sampling rate of the original Egotracks dataset is 5 FPS.
From Table~\ref{table:training-dataset}, we can observe that our \trainingdataset{} is characterized by:
\begin{itemize}
    \item \textbf{Shorter} total duration: On average, the referred objects show in the video for only \textbf{21.56}\% time, compared to over 75\% time in previous datasets.
    \item \textbf{Sparser} continuous trajectories: On average, the length of continuous trajectory (\ie, consecutive appearance) is only \textbf{1.33}\% of the whole video, while in the exocentric video, the object trajectory length is over \textbf{65}\%.
    Also, the disappearance of target objects is about six times longer than their appearance on average (\textbf{655.82\%} Disappear Ratio), while in exocentric videos, the disappearance duration is much shorter (under 21\%).
    % Also, the average time of each consecutive disappearance is about six times longer than that of the appearance (\textbf{655.82\%} Disappear Ratio(\%)). In contrast, in exocentric data, the disappearance is much shorter.
    %
    \item \textbf{Smaller} object size: The average mask area is only \textbf{1.20}\% of the whole frame, while objects in the previous benchmarks are five times larger (over 5\%).
    \item \textbf{Larger} positional shift: The average IoU of masks on the adjacent frames is only \textbf{14.96}\%, while in the exocentric video, the IoU value is over 50\%. It is caused by the large and frequent camera movement in egocentric videos.
\end{itemize}

\smallskip 
\noindent \textbf{Evaluation benchmark curation.} Our benchmark \textbf{\textit{\benchmark{}}} includes 315 videos ranging from 5 seconds to 16 minutes, categorized into short-term (under 1 minute), medium-term (between 1 minute and 3 minutes), and long-term (over 3 minutes). 
It contains 700 expressions in total with an average of 15 words.
All samples are manually refined and verified by using the semi-automatic labeling tool~\cite{ISAT_with_segment_anything} that supports SAM2 tracking. 

\begin{itemize}
    \item \benchmark{}-Short: The short videos and the corresponding expressions are sampled from RefEgo~\cite{DBLP:conf/iccv/KuritaKO23}. The annotators manually label the object masks and refine the expressions to make them more precise. 
    \item \benchmark{}-Long: The long videos are sampled from the validation set of Egotracks ~\cite{DBLP:conf/nips/TangLGF023}. We use the above pipeline to generate the masks and expressions for human annotators to refine, which can largely reduce the annotation time.
    \item \benchmark{}-Medium: The medium videos are extracted from the above annotated long videos. For each object in the long video, we randomly extract two different clip segments that contain the target object.
\end{itemize}

The statistics of \benchmark{} are shown in Table~\ref{table:benchmark-test}, alongside the comparison with existing exocentric benchmarks. Same as our training dataset, our egocentric benchmark is characterized by \textbf{shorter} total duration, \textbf{sparser} continuous trajectories,\textbf{smaller} object size, and  \textbf{larger} positional shift. 
%
% Note that videos of \benchmark{}-Short also have 12 seconds, so its object total duration and sparsity of continuous trajectories are not apparent.

\section{Experiments}
In this section, we first introduce the evaluation dataset, evaluation metrics, the baseline models, and implementation details. Then we present our results, including the results of our egocentric benchmark, exocentric benchmarks, detailed analysis, and visualizations.

\subsection{Setup}

\smallskip \noindent 
\textbf{Dataset.} 
We evaluate pixel-level spatiotemporal video grounding on one egocentric benchmark (our proposed \benchmark{}) and three exocentric benchmarks (Ref-Davis~\cite{DBLP:conf/accv/KhorevaRS18}, Mevis~\cite{DBLP:conf/iccv/Ding0H0L23}, and ReasonVOS~\cite{DBLP:conf/nips/Bai0MWGClZS24}).
For model fine-tuning, apart from our proposed dataset \trainingdataset{}, we also use a mixture of image-based and video-based segmentation datasets~\cite{DBLP:conf/nips/Bai0MWGClZS24,DBLP:conf/cvpr/ZhouZPFB017,DBLP:conf/cvpr/CaesarUF18,DBLP:conf/cvpr/RamanathanKPWZG23,DBLP:conf/cvpr/ChenMLFUY14,DBLP:conf/emnlp/KazemzadehOMB14,DBLP:conf/cvpr/MaoHTCY016,DBLP:conf/eccv/SeoLH20,DBLP:conf/iccv/Ding0H0L23,DBLP:conf/accv/KhorevaRS18,DBLP:journals/corr/abs-1809-03327,DBLP:conf/eccv/YanWYJHKXG24}, following the previous works~\cite{DBLP:conf/nips/Bai0MWGClZS24,DBLP:journals/corr/abs-2501-04001}.
These segmentation datasets focus on exocentric videos, while our training dataset is tailored for egocentric videos and has distinct characteristics.

\smallskip 
\noindent \textbf{Metrics.} For fine-grained spatiotemporal grounding on egocentric videos, we adopt four metrics. 
We refer ``target frame'' as the video frame where the target (ground-truth) entity presents, ``background frame'' as the video frames that do not have the target entity,  and ``predicted frame'' as the video frame with predicted masks from the models.

\begin{itemize}
    \item $\TRecall{}$: It is defined as the proportion of predicted frames among target frames. It evaluates the temporal grounding performance of models.
    \item $\IouAll{}$: It is calculated by the average of Intersection-over-Unions~(IoUs) value among all the video frames, which is also called region similarity  ($\mathcal{J}$) and commonly adopted in the previous works~\cite{DBLP:journals/corr/Pont-TusetPCASG17, DBLP:conf/nips/Bai0MWGClZS24}.
    \item $\IouGold{}$: It is defined by the average of IoU among target frames. This metric only focuses on the prediction results over target masks and \textit{ignores} the predictions on the background frames.
    \item $\IouGoldPred{}$: It is defined by the average of IoUs among all target frames and predicted frames. It excludes the frames where the ground-truth mask and predicted mask do not exist.
    This metric is more challenging as it \textit{penalizes} the hallucinated predictions on the background frames.
\end{itemize}

\smallskip \noindent \textbf{Baselines.} We choose one pipeline tracker and two end-to-end VideoLLMs as our baselines.

\begin{itemize}
\item Grounded-SAM2~\cite{ren2024grounded}, a pipeline tracker. It first uses GroundingDino~\cite{liu2023grounding} to perform image-based object detection and output a bounding box. Then, the box is viewed as the box prompt for the SAM2 model to get the segmentation masks.
For the first stage, we apply GroundingDino on all the frames and select the frame whose generated bounding box has the highest detection confidence score as the initial frame for SAM2 tracking.
\item Sa2VA~\cite{DBLP:journals/corr/abs-2501-04001}, an open-sourced VideoLLM which links LLaVA and SAM2 to perform video grounding. It first generates the mask for a few key frames, and then uses SAM2 to perform video propagation based on these frames. 
During inference, it uses the first five frames as the key frames. We test its 4B and 26B versions in our evaluation. 

\item VideoLISA~\cite{DBLP:conf/nips/Bai0MWGClZS24}, an open-source VideoLLM which uses ``[SEG]'' token to link LLaVA-Phi-3-V~\cite{rasheedllava++} and SAM. It performs independent image segmentation on the frames using the same [SEG] token. Its model size is 3.8B.

\end{itemize}

\smallskip \noindent \textbf{Fine-tuning Implementation.} We build two fine-tuned VideoLLMs, which are shown as follows,
\begin{itemize}
    \item Sa2VA-4B (+FT): We fine-tune Sa2VA-4B~\cite{DBLP:journals/corr/abs-2501-04001} on our proposed egocentric dataset \trainingdataset{}, alongside three video segmentation datasets~(Mevis~\cite{DBLP:conf/iccv/Ding0H0L23}, ReasonVOS~\cite{DBLP:conf/nips/Bai0MWGClZS24}, 
    Ref-YouTube-VOS~\cite{DBLP:conf/eccv/SeoLH20}) that the original model is trained on. 
    The fine-tuning is deployed on 8 NVIDIA 80G A100 GPUs for about 10 hours, with AdamW~\cite{DBLP:conf/iclr/LoshchilovH19} optimizer and learning rate as 4e-6. The batch size is set to 16.
    \item VideoLISA-3.8B (+FT): 
    We fine-tune VideoLISA~\cite{DBLP:conf/nips/Bai0MWGClZS24} using 4 NVIDIA 80G A100 GPUs based on DeepSpeed~\cite{DBLP:conf/kdd/RasleyRRH20} for a total of 20 epochs, with AdamW~\cite{DBLP:conf/iclr/LoshchilovH19} optimizer and learning rate as 3e-5.
    The fine-tuning data contains two parts, with 80\% sampled from our proposed egocentric dataset \trainingdataset{} and 20\% sampled from a suite of image and video segmentation datasets that the original model is trained on.
    The batch size is set to 16, and the number of steps for each epoch is set to 500.
    The whole process takes about 12 hours.
    %  resulting in a total of 8,000 data samples per epoch. The whole process takes about 12 hours.
\end{itemize}

\subsection{Main Results}

% \usepackage{multirow}
% \usepackage{graphicx}
\begin{table*}[t!]
\centering
\resizebox{0.95\textwidth}{!}{%
\begin{tabular}{c|cccl|cccl|cccl}
\toprule

\multirow{2}{*}{\textbf{Methods}} &
  \multicolumn{4}{c|}{\textbf{Short}} &
  \multicolumn{4}{c|}{\textbf{Medium}} &
  \multicolumn{4}{c}{\textbf{Long}} \\ 
  \cmidrule{2-13}
 &
  $\TRecall{}$ &
  $\IouAll{}$ &
  $\IouGold{}$ &
  $\IouGoldPred{}$ &
  $\TRecall{}$ &
  $\IouAll{}$ &
  $\IouGold{}$ &
  $\IouGoldPred{}$ &
  $\TRecall{}$ &
  $\IouAll{}$ &
  $\IouGold{}$ &
  $\IouGoldPred{}$  \\ \midrule

\multicolumn{13}{c}{\textit{\textbf{Pipeline Baseline}}} \\ 
\midrule
Grounded-SAM2  & 91.31 & 54.75 & 51.00 & \textbf{49.95} 
               & 65.85 & 54.35 & 28.23 & \textbf{25.73} 
               & 61.44 & 61.54 & 27.36 & \textbf{24.80}  \\ 
\midrule
\multicolumn{13}{c}{\textit{\textbf{End2End Open-Source VideoLLMs}}} \\ 
\midrule
Sa2VA-26B & 70.08 & 48.23 & 39.20 & 37.30 
          & 44.15 & 73.06 & 28.83 & 25.83 
          & 28.53 & 72.01 & 15.45 & 12.96  \\ 

Sa2VA-4B & 69.33 & 40.41 & 31.01 & 29.00 
         & 36.83 & 66.10 & 18.68 & 17.02 
         & 21.55 & 68.67 & 8.68  & 8.11 \\ 

Sa2VA-4B  (+FT)  & 72.62 & 41.70 & 32.92 & 30.97 (+\textbf{1.97}) 
                 & 39.27 & 67.42 & 20.12 & 18.52 (+\textbf{1.50}) 
                 & 21.60 & 69.50 & 9.14  & 8.24 (+\textbf{0.13}) \\ 

\midrule

\midrule  
VideoLISA-3.8B   & 98.37 & 18.14 & 20.94 & 17.85 
                 & 96.99 & 9.28  & 11.87 & 6.48  
                 & 96.13 & 8.48  & 12.11 & 5.15\\ 
% \midrule
 
VideoLISA-3.8B (+FT) & 97.95 & 23.98 & 27.33 & 23.36 (+\textbf{5.51}) 
                     & 95.52 & 14.05 & 17.28 & 9.98 (+\textbf{3.50})  
                     & 95.33 & 12.03 & 14.82 & 7.16 (+\textbf{2.01})\\ 

\bottomrule
\end{tabular}%
}
\vspace{-3pt}
\caption{Experimental results on our egocentric benchmark \benchmark{}. FT means the fine-tuning on our training dataset \trainingdataset{}.}
\label{table:main-results-on-benchmark}
\end{table*}
% \usepackage{multirow}
% \usepackage{graphicx}
\begin{table*}[t!]
\centering
\resizebox{0.95\textwidth}{!}{%
\begin{tabular}{c|cccl|cccl|cccl}
\toprule
\multirow{2}{*}{\textbf{Methods}} &
  \multicolumn{4}{c|}{\textbf{Ref-Davis}} &
  \multicolumn{4}{c|}{\textbf{Mevis}} &
  \multicolumn{4}{c}{\textbf{ReasonVOS}} \\ \cmidrule{2-13} 
 &
  $\TRecall{}$ & $\IouAll{}$ & $\IouGold{}$ & $\IouGoldPred{}$ &
  $\TRecall{}$ & $\IouAll{}$ & $\IouGold{}$ & $\IouGoldPred{}$ &
  $\TRecall{}$ & $\IouAll{}$ & $\IouGold{}$ & $\IouGoldPred{}$ \\ \midrule

\multicolumn{13}{c}{\textit{\textbf{Pipeline Baseline}}} \\ 
\midrule

Grounded-SAM2 & 98.48 & 62.74 & 62.41 & 62.39 
              & 97.27 & 41.07 & 40.69 & 40.65 
              & 0.00  & 9.91  & 0.00  & 0.00 \\  \midrule

\multicolumn{13}{c}{\textit{\textbf{End2End Open-Source VideoLLMs}}} \\ 
\midrule
Sa2VA-26B  & 96.90 & 73.77 & 74.12 & 73.58 
           & 95.98 & 56.38 & 55.24 & 54.80 
           & 79.84 & 57.34 & 53.21 & 52.47 \\ 
Sa2VA-4B & 96.21 & 69.87 & 70.22 & 69.75 
         & 96.54 & 51.24 & 50.41 & 50.01 
         & 77.69 & 47.50 & 43.35 & 42.35 \\ 
Sa2VA-4B~(+FT) & 96.22 & 70.09 & 70.45 & 69.97 (+\textbf{0.22}) 
               & 96.21 & 57.45 & 56.05 & 55.55 (+\textbf{5.54}) 
               & 79.10 & 50.11 & 46.34 & 45.54 (+\textbf{3.19}) \\ 
\midrule

\midrule
VideoLISA-3.8B & 99.99 & 65.84 & 66.26 & 65.82 
               & 99.75 & 49.36 & 51.18 & 49.20 
               & 99.66 & 42.43 & 45.48 & 42.41 \\ 
% \midrule

VideoLISA-3.8B (+FT) & 99.96 & 65.60 & 66.01 & 65.60 (-\textbf{0.22}) 
                     & 99.54 & 49.46 & 51.31 & 49.20 (-\textbf{0.00}) 
                     & 99.09 & 44.27 & 47.13 & 44.18 (+\textbf{1.77}) \\ \bottomrule
\end{tabular}%
}
\vspace{-3pt}
\caption{Experimental results on existing exocentric benchmarks. FT means the fine-tuning on our training dataset \trainingdataset{}.}
\label{table:main-result-on-exocentric-benchmark}
\vspace{-3pt}
\end{table*}

We conduct experiments with various grounding methods on our proposed egocentric benchmark \benchmark{} (Table~\ref{table:main-results-on-benchmark}) and three existing exocentric benchmarks (Table~\ref{table:main-result-on-exocentric-benchmark}).

\smallskip \noindent \textbf{Existing models perform poorly on egocentric videos.} 
As Table~\ref{table:main-results-on-benchmark} shows, all grounding methods fail to achieve satisfactory performance on our benchmark.
For the least challenging subset of our benchmark (\benchmark{}-Short), the best $\IouGoldPred{}$ score is under 50\%. 
And for \benchmark{}-Medium and \benchmark{}-Long, the challenging subsets with longer video lengths, the performance drops largely for all models (\ie, less than 30\% $\IouGoldPred{}$ score). 
Our Benchmark annotations have been manually verified to minimize the bias from annotation tools. Even though some baselines adopt the SAM2 model, which is also used in our annotation pipeline, they still fail to perform well on our benchmark (\ie, Grounded-SAM2 only achieves 49.95\%, 25.73\%, and 24.80\% on Short, Medium, and Long, respectively).
These results reveal that our pixel-level egocentric grounding benchmark is challenging, and there is a significant gap between egocentric and exocentric videos. % on which the models are trained.

\smallskip 
\noindent \textbf{Our training dataset is beneficial.}
The models fine-tuned on our training set achieve large improvements on our benchmark, while maintaining their grounding ability in exocentric benchmarks.
As in Table~\ref{table:main-results-on-benchmark}, VideoLISA-3.8B~(+FT) achieves a noticeable performance improvement on all subsets, with an average relative increase of 41.30\% and absolute increase of 3.67\%. Similarly, Sa2VA-4B~(+FT) achieves an average relative increase of 5.74\% and absolute increase of 1.20\%.
On the other hand, when evaluated on exocentric benchmarks, the fine-tuned models achieve comparable or even better results than the original pre-trained models. VideoLISA-3.8B~(+FT) achieves a 1.77\% performance improvement in the ReasonVOS benchmark and drops slightly in Ref-Davis (-0.22\%). Sa2VA-4B~(+FT) surpasses its baseline on all benchmarks.
These results indicate that our training dataset is complementary to previous exocentric datasets.
Overall, these experiments verify the effectiveness of our proposed training dataset \trainingdataset {} for improving fine-grained egocentric spatiotemporal grounding, providing a unique resource for future video foundation models.

\smallskip \noindent \textbf{Our defined metrics can better reflect the grounding capabilities on both egocentric and exocentric videos.}
The commonly-used metric $\IouAll{}$ (also known as $\mathcal{J}$) is not proper for our egocentric benchmark,
especially in \benchmark{}-Medium and \benchmark{}-Long.
This arises from the large portion of background frames. 
% Explanation
The $\IouAll{}$ metric considers all frames and thus is dominated by the background frames, failing to reflect the model performance at target frames.
In contrast, our proposed metrics $\IouGold{}$/$\IouGoldPred{}$ will ignore/penalize the predictions on background frames, thereby better reflecting model performance.
%% example
Take Sa2VA in Table~\ref{table:main-results-on-benchmark} as an example. We can observe that from short to long, the model retrieves fewer target frames (\textbf{$\TRecall{}$} is small), while the $\IouAll{}$ score gets higher. 
In comparison, our defined metrics are more stable and align with the subset difficulty (\ie, $\IouGold{}$/$\IouGoldPred{}$ drops as $\TRecall{}$ decreases). 
% exocentric
Meanwhile, exocentric benchmarks include fewer background frames, and thus, our proposed metrics degenerate to the old metric. As shown in Table~\ref{table:main-result-on-exocentric-benchmark}, our proposed metrics are consistent with $\IouAll{}$.
Therefore, our defined metrics well reflect grounding abilities on both ego and exo videos.

\smallskip \noindent \textbf{Model design tailored for egocentric videos is lacking.} 
The comparison between VideoLISA and Sa2VA suggests that transferring video segmentation ability from the SAM2 model is better than performing independent image segmentation on individual frames.
We also compare Sa2VA and Grounded-SAM2, both of which are SAM2-based methods. Their major difference lies in their architectures: Sa2VA is an end-to-end model while Grounded-SAM2 follows a pipeline design. We find that Sa2VA shows better text query understanding than the pipeline model, as reflected by the results on ReasonVOS, and performs better on exocentric benchmarks (Table~\ref{table:main-result-on-exocentric-benchmark}). 
The main reason is that Grounded-SAM2 relies on the boxes predicted from GroundingDino, which fails to fully understand the expressions. It thus achieves low grounding scores due to the wrong boxes and the error propagation. In contrast, Sa2VA can take advantage of the reasoning capabilities of LLMs to better understand the queries.
However, on \benchmark{}, even the largest Sa2VA model (26B) achieves inferior performance to Grounded-SAM2. These observations indicate that the existing modeling fails to explore the full potential of the pre-trained grounding models on egocentric videos.

\subsection{Analysis}

\begin{table}[ht]
\centering
\resizebox{1.0\columnwidth}{!}{%
\begin{tabular}{c|c|cl|cl}
\toprule
\multirow{2}{*}{\textbf{\benchmark{}}} & \textbf{Ground with highest} & \multicolumn{2}{c}{\textbf{Detection}} & \multicolumn{2}{|c}{\textbf{\textit{ST}-Grounding}}  \\
\cmidrule{3-6}
 &   \textbf{detection confidence} & \textit{Accuracy} & \textit{IoU} & $\TRecall{}$ & $\IouGoldPred{}$ \\
 \midrule
\multirow{2}{*}{Short}  &  \cmark & 87.00 & 52.97 & 91.31 & 49.95 \\
                        &  \xmark & 82.75 & 42.49 (-\textbf{10.48}) & 87.64 & 40.42 (-\textbf{9.53})\\
 \midrule
\multirow{2}{*}{Medium} &  \cmark & 51.50 & 31.56 & 65.85 & 25.73  \\
                        &  \xmark & 67.00 & 18.24 (-\textbf{13.32}) & 65.38 & 15.11 (-\textbf{10.62}) \\
 \midrule
\multirow{2}{*}{Long}   &  \cmark & 47.00 & 32.74 & 61.44 & 24.80  \\
                        &  \xmark & 34.00 & 14.20 (-\textbf{18.54}) & 54.36 & 11.65 (-\textbf{13.15}) \\
\bottomrule
\end{tabular}%
}
\vspace{-2pt}
\caption{Experimental results for different initialization states of SAM2 in Grounded-SAM2, where ST means spatiotemporal, and the initialization state refers to the box prompt detected by GroundingDino.}
\label{table:groundedsam2-on-benchmark}
\vspace{-5pt}
\end{table}

\begin{table}[ht]
\centering
\resizebox{\columnwidth}{!}{%

\begin{tabular}{c|c|c|cc|cc}
\toprule
\multirow{2}{*}{\textbf{Type}} & 
\textbf{Valid Key}   & 
\textbf{\#Test}  & 
\multicolumn{2}{c|}{\textbf{Sa2VA-26B}} & 
\multicolumn{2}{c}{\textbf{Sa2VA-4B}} \\
            &    \textbf{Frames}   &   \textbf{sample}  &{$\TRecall{}$} & {$\IouGoldPred{}$} & {$\TRecall{}$} & {$\IouGoldPred{}$}  \\

\midrule

\multirow{2}{*}{Short} & \cmark & 394 & 70.93 & 37.87 & 70.38 & 29.44   \\
                       & \xmark & 6   & 14.58 & 0.00  & 0.00  & 0.00   \\
\midrule
\multirow{2}{*}{Medium}   & \cmark & 150 & 54.64 & 34.15 & 42.96 & 22.44  \\
                       & \xmark & 50  & 12.67 & 0.86  & 18.42 & 0.78          \\
\midrule
\multirow{2}{*}{Long}  & \cmark & 48  & 46.24 & 25.59 & 32.73 & 16.42     \\
                       & \xmark & 52  & 12.18 & 1.29  & 11.23 & 0.44 \\

\bottomrule
\end{tabular}
}
\vspace{-2pt}
\caption{Experimental results for different initialization states of SAM2 in Sa2VA, where the initialization state means the mask generated for key frames. The ``Valid Key Frames'' means the first five frames contain targeted frames.}
\label{table:sa2va_5frames}
\vspace{-5pt}
\end{table}

\begin{table}[ht]
\centering
\resizebox{\columnwidth}{!}{%
\begin{tabular}{ccccc}
\toprule
  \textbf{Methods}            & \textbf{VideoLLM} & \textbf{SAM-2} & \textbf{Variation} & \textbf{Speed (FPS)} \\ \midrule
\multirow{2}{*}{Grounded-SAM2} &   \multirow{2}{*}{\xmark}    &  \multirow{2}{*}{\cmark}  & \textit{w.  highest conf.}  &   3.17 \\ 
              &       &   & \textit{w.o highest conf.}  &   7.14 \\ \midrule

% \multirow{2}{*}{Sa2VA}    &      \multirow{2}{*}{\cmark}    &  \multirow{2}{*}{\cmark}  & 4B  &   6.47 \\ 
%               &      &   & 8B  &   5.26 \\ \midrule
Sa2VA-4B        & \cmark & \cmark & - & 6.47 \\ \midrule
VideoLISA-3.8B     &   \cmark    &  \xmark  & -   &   0.42 \\ \bottomrule
\end{tabular}%
}
\vspace{-2pt}
\caption{Comparison of grounding speed. For Grounded-SAM2, the ``\textit{w. highest conf.}'' means it performs grounding with the highest detection confidence score and ``\textit{w.o highest conf.}'' means the naive variant that uses the first detected object as the box prompt. }
\label{table:grounding-speed}
\vspace{-8pt}
\end{table}

We conduct further analysis to evaluate the modeling design and reveal corresponding challenges.

\smallskip 
\noindent \textbf{Effects of Initialization State of SAM2.} From Table~\ref{table:main-results-on-benchmark}, the SAM2-based models (Grounded-SAM2 and Sa2VA) perform better than the SAM-based model (VideoLISA) as they can utilize information from the context frames.
However, for these SAM2-based models, the initialization of the inference state is of vital importance and influences the performance a lot due to the information propagation. 
We conduct experiments with the SAM2-based methods (shown in Table~\ref{table:groundedsam2-on-benchmark} and Table~\ref{table:sa2va_5frames}). 
(1) For Grounded-SAM2, its inference state is initialized by the detected bounding box from GroundingDino.
Thus, we implement a naive variant that, instead of using the detected bounding box with the highest confidence for grounding, we simply use the first detected bounding box as the prompt, regardless of its confidence score. 
The results are shown in Table~\ref{table:groundedsam2-on-benchmark}. As the detection performance drops in the naive variant, the grounding performance also drops, essentially at the average of 11.1\%. We also witness the same situation on exocentric video grounding tasks, with the average of 4.65\% score reduced. 
(2) For Sa2VA, the inference state of SAM2 is initialized by the generated mask on the input key frames. As we mentioned before, the Sa2VA model usually takes key frames as the first five frames of the input video. When the referred entity does not show in these frames, the key frames are viewed as invalid.
The performance in Table~\ref{table:sa2va_5frames} shows that when the model fails to initialize the SAM2 properly, the overall grounding performance will suffer from rapid deduction and drop to nearly 0.00\%. 
These results reveal that \textit{Utilizing SAM2 is beneficial for grounding, yet a proper initialization state is critical to fully unleash its capability}.

The initialization state poses a challenge for modeling design. On one hand, the limited input tokens of the current grounding models restrict them from taking more frames to provide initial information for SAM2. On the other hand, the ego videos are usually long, and the target objects have shorter total durations and sparser continuous trajectories. So the sampled frames are more likely to be the background frames. 
Possible solutions could be: (1) enhancing long video understanding capabilities, and (2) optimizing frame selection to capture the target entities.

\smallskip 
\noindent \textbf{Inference Speed.} 
We measure the inference speed of each method in
Table~\ref{table:grounding-speed}. The results demonstrate that performing image-level segmentation is inefficient (\eg, only \textbf{0.42} FPS), while SAM2-based methods with video-level segmentation generally achieve higher speed (\eg, at least \textbf{3.17} FPS).

\smallskip 
% \noindent \textbf{Modeling Insights.}
\noindent
The above analysis suggests that:
\infobox{Utilizing SAM2 for spatiotemporal grounding offers advantages in both effectiveness and efficiency; however, careful initialization is crucial to fully unleash its pre-trained capability.}

% More analysis can be found in the Appendix.

% \smallskip 
% \noindent \textbf{Modeling Insights.} Based on the above results and analysis, we provide the following insights about modeling for spatiotemporal video grounding.
% \begin{itemize}
%     \item Adopting SAM2 for spatiotemporal grounding is beneficial for spatiotemporal grounding on both \textit{effectivity} (shown in Table~\ref{table:main-results-on-benchmark}) and \textit{efficiency} (shown in Table~\ref{table:grounding-speed}), but its initialization matters.
%     \item Improving long video understanding capabilities is needed. Most VideoLLMs use frame sampling strategies, resulting in inevitable information loss.  For videos that have a longer length and the objects have a shorter total duration, chances are that the target objects are not in the sampled frames, which may cause grounding errors. 
    
% \end{itemize}

\subsection{Visualization examples}
\begin{figure}[t]
\begin{center}
\includegraphics[width=0.80\linewidth]{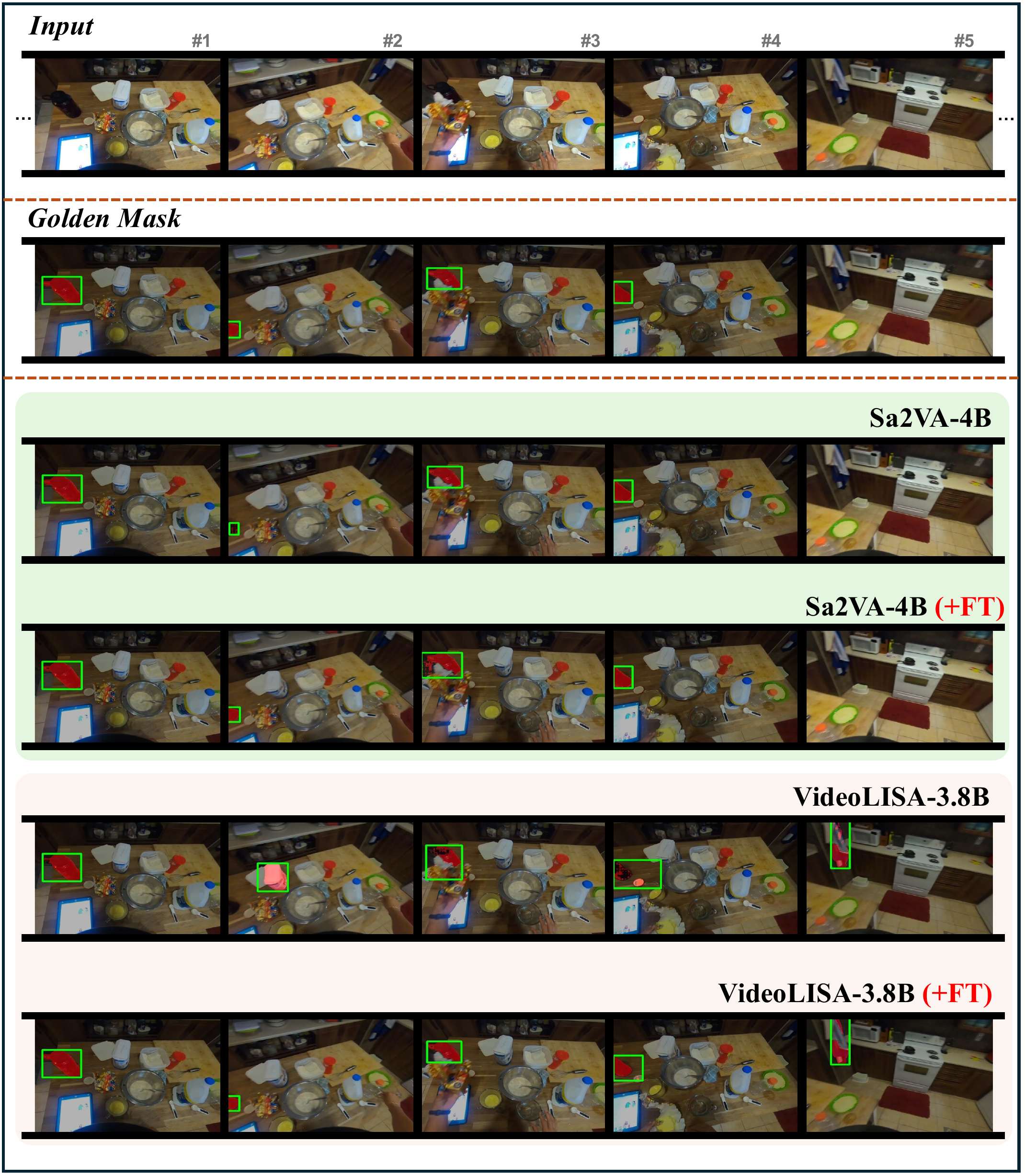}
\end{center}
\vspace{-15pt}
   \caption{Visualization of one example from \benchmark{}-Short with sampled frames. The language query is ``black container bottle on the left side of a wooden table behind computer tablet''. The fine-tuned models perform better than their zero-shot counterparts.
   } 
   \vspace{-10pt}
\label{fig:short_visualization}
\end{figure}

\begin{figure}[t]
\begin{center}
\includegraphics[width=0.80\linewidth]{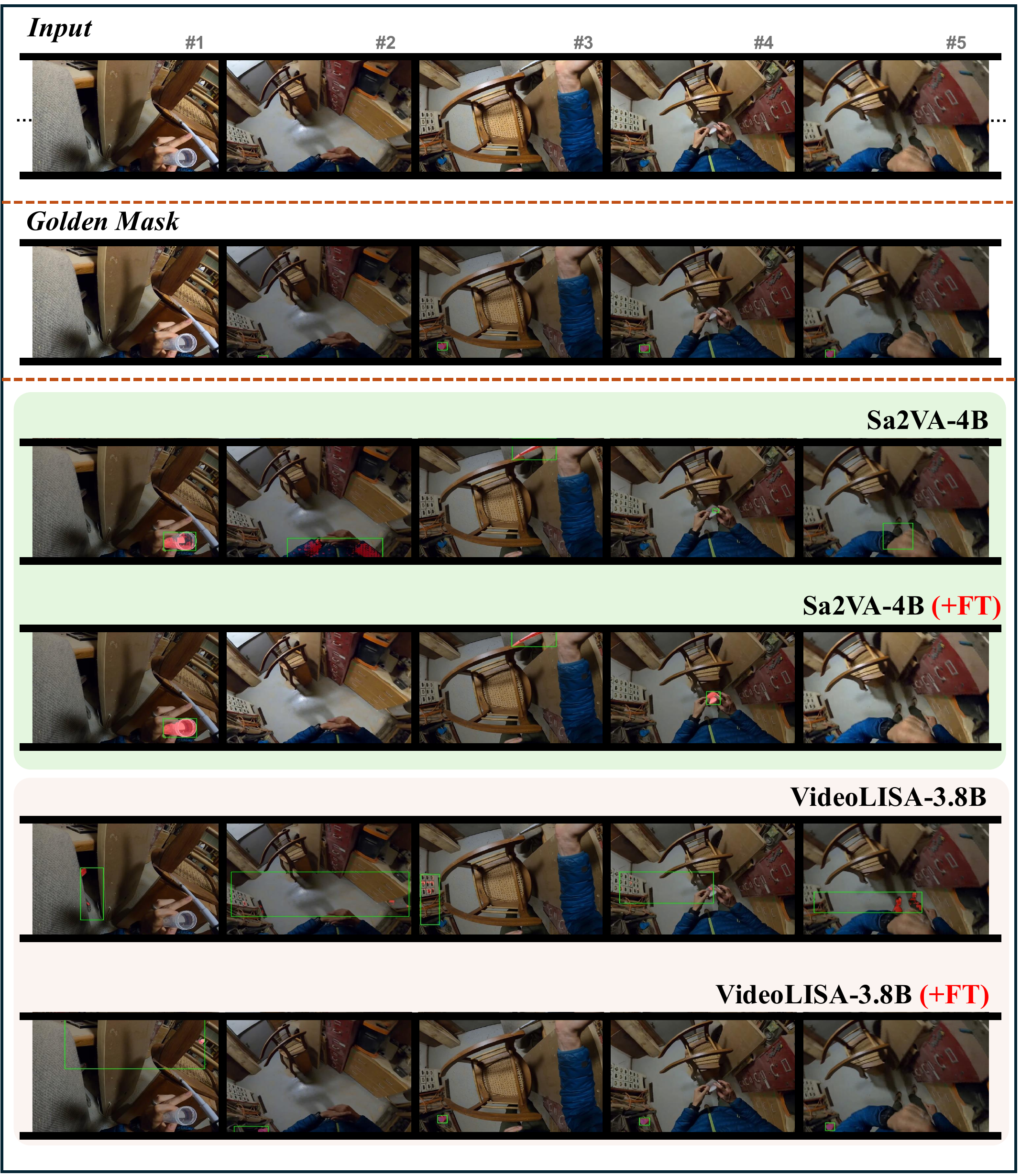}
\end{center}
\vspace{-15pt}
   \caption{Visualization of one example from \benchmark{}-Long with sampled frames. The language query is ``small blue cylindrical container near the floor''. The small target poses a challenge to existing methods.
   } 
   \vspace{-10pt}
\label{fig:long_visualization}
\end{figure}
We visualize a few examples from ~\benchmark{} and compare the results from different models 
% (Figure~\ref{fig:short_visualization},\ref{fig:mid_visualization}, and \ref{fig:long_visualization}).
(Figure~\ref{fig:short_visualization} and Figure~\ref{fig:long_visualization}).

\smallskip 
\noindent \textbf{Small object size poses a challenge to existing methods.} 
As Figure~\ref{fig:long_visualization} shows, the object is very small and usually near the edge of the view, which challenges existing methods. Both Sa2VA-4B and VideoLISA fail to ground the target.

\smallskip 
\noindent \textbf{Our training dataset is beneficial.} 
The fine-tuned VideoLISA model can correct grounding errors compared with its zero-shot counterpart. 
In Figure~\ref{fig:short_visualization}, where the pre-trained model ignores the less salient target near the edge and locates the wrong object (\#2 frame), the fine-tuned model identifies the object correctly. 
Also in Figure~\ref{fig:long_visualization}, the fine-tuned model successfully locates the small object (\#3 - \#5 frame), which the pre-trained model fails to ground. 
Similarly, the fine-tuned Sa2VA-4B performs better than its original version and outputs more precise masks (\ie, \#2 frame in Figure~\ref{fig:short_visualization}).
These results indicate that our training data can facilitate the models in addressing the challenges in egocentric videos.
% In Figure~\ref{fig:mid_visualization}, even though our fine-tuned model grounds the wrong object at the beginning, it performs best among the four methods. Sa2VA and the original VideoLISA model ground the wrong object, and the Grounded-SAM2 fails to generate masks. 
% In our fine-tuned model successfully locates on the small object, while the pre-trained model is unable to ground.

\smallskip 
\noindent \textbf{Ultilizing SAM2 provides benefits.} 
% In egocentric videos that feature frequent and rapid movement, the target entities have larger positional shifts. 
From Figure~\ref{fig:short_visualization}, we notice inconsistent segmentation from two VideoLISA models~(\#5 frame), while other SAM2-based models can correctly segment the target across video frames without hallucinations. This verifies the effectiveness of the SAM2 model as it can utilize information from the context frames through its memory module. 
% In contrast, the VideoLISA adopts the SAM model and performs independent image segmentation on video frames, resulting in segmentation inconsistency.  

\section{Conclusion}

To address the absence of benchmarks for egocentric spatiotemporal grounding, we introduce \benchmark{}, a benchmark with pixel-level annotations, alongside a large-scale training dataset \trainingdataset{}. 
Through extensive experiments, we demonstrate that existing state-of-the-art models struggle with egocentric videos but significantly improve when fine-tuned on our training data, while retaining performance on exocentric benchmarks. These findings, along with our in-depth analysis on ego-exo video gaps, highlight the importance of egocentric training data and suggest that our benchmark can serve as a valuable resource for future research in video grounding and embodied AI.

% In this work, we address the lack of benchmarks for egocentric spatiotemporal grounding by introducing \benchmark{}, a benchmark with pixel-level annotations, alongside a large-scale training dataset \trainingdataset{}. Through extensive experiments, we demonstrate that existing SOTA models struggle with egocentric videos but significantly improve when fine-tuned on our dataset, while retaining performance on exocentric benchmarks. These findings, along with our in-depth analysis on ego-exo video gaps, highlight the importance of egocentric-specific training data and suggest that our benchmark can serve as a valuable resource for future research in video grounding and embodied AI.

\section*{Acknowledgements}
This work was supported by National Key R\&D Program of China (Project No. 2022ZD0161200, 2022ZD0161201). This work has also been supported by Hong Kong Research Grant Council - Early Career Scheme (Grant No. 24200223) as well as partially supported by Hong Kong Innovation and Technology Commission Project No. ITS/228/22FP.

{
    \small
    \bibliographystyle{ieeenat_fullname}
    \bibliography{main}

@inproceedings{DBLP:conf/cvpr/GraumanWBCFGH0L22,
  author       = {Kristen Grauman and
                  Andrew Westbury and
                  Eugene Byrne and
                  Zachary Chavis and
                  Antonino Furnari and
                  Rohit Girdhar and
                  Jackson Hamburger and
                  Hao Jiang and
                  Miao Liu and
                  Xingyu Liu and
                  Miguel Martin and
                  Tushar Nagarajan and
                  Ilija Radosavovic and
                  Santhosh Kumar Ramakrishnan and
                  Fiona Ryan and
                  Jayant Sharma and
                  Michael Wray and
                  Mengmeng Xu and
                  Eric Zhongcong Xu and
                  Chen Zhao and
                  Siddhant Bansal and
                  Dhruv Batra and
                  Vincent Cartillier and
                  Sean Crane and
                  Tien Do and
                  Morrie Doulaty and
                  Akshay Erapalli and
                  Christoph Feichtenhofer and
                  Adriano Fragomeni and
                  Qichen Fu and
                  Abrham Gebreselasie and
                  Cristina Gonz{\'{a}}lez and
                  James Hillis and
                  Xuhua Huang and
                  Yifei Huang and
                  Wenqi Jia and
                  Weslie Khoo and
                  J{\'{a}}chym Kol{\'{a}}r and
                  Satwik Kottur and
                  Anurag Kumar and
                  Federico Landini and
                  Chao Li and
                  Yanghao Li and
                  Zhenqiang Li and
                  Karttikeya Mangalam and
                  Raghava Modhugu and
                  Jonathan Munro and
                  Tullie Murrell and
                  Takumi Nishiyasu and
                  Will Price and
                  Paola Ruiz Puentes and
                  Merey Ramazanova and
                  Leda Sari and
                  Kiran Somasundaram and
                  Audrey Southerland and
                  Yusuke Sugano and
                  Ruijie Tao and
                  Minh Vo and
                  Yuchen Wang and
                  Xindi Wu and
                  Takuma Yagi and
                  Ziwei Zhao and
                  Yunyi Zhu and
                  Pablo Arbel{\'{a}}ez and
                  David Crandall and
                  Dima Damen and
                  Giovanni Maria Farinella and
                  Christian Fuegen and
                  Bernard Ghanem and
                  Vamsi Krishna Ithapu and
                  C. V. Jawahar and
                  Hanbyul Joo and
                  Kris Kitani and
                  Haizhou Li and
                  Richard A. Newcombe and
                  Aude Oliva and
                  Hyun Soo Park and
                  James M. Rehg and
                  Yoichi Sato and
                  Jianbo Shi and
                  Mike Zheng Shou and
                  Antonio Torralba and
                  Lorenzo Torresani and
                  Mingfei Yan and
                  Jitendra Malik},
  title        = {Ego4D: Around the World in 3, 000 Hours of Egocentric Video},
  booktitle    = {{CVPR}},
  pages        = {18973--18990},
  publisher    = {{IEEE}},
  year         = {2022}
}

@inproceedings{DBLP:conf/nips/TangLGF023,
  author       = {Hao Tang and
                  Kevin J. Liang and
                  Kristen Grauman and
                  Matt Feiszli and
                  Weiyao Wang},
  title        = {EgoTracks: {A} Long-term Egocentric Visual Object Tracking Dataset},
  booktitle    = {NeurIPS},
  year         = {2023}
}

@inproceedings{DBLP:conf/iccv/KuritaKO23,
  author       = {Shuhei Kurita and
                  Naoki Katsura and
                  Eri Onami},
  title        = {RefEgo: Referring Expression Comprehension Dataset from First-Person
                  Perception of Ego4D},
  booktitle    = {{ICCV}},
  pages        = {15168--15178},
  publisher    = {{IEEE}},
  year         = {2023}
}

@inproceedings{DBLP:conf/iccv/Ding0H0L23,
  author       = {Henghui Ding and
                  Chang Liu and
                  Shuting He and
                  Xudong Jiang and
                  Chen Change Loy},
  title        = {MeViS: {A} Large-scale Benchmark for Video Segmentation with Motion
                  Expressions},
  booktitle    = {{ICCV}},
  pages        = {2694--2703},
  publisher    = {{IEEE}},
  year         = {2023}
}

@article{DBLP:journals/corr/Pont-TusetPCASG17,
  author       = {Jordi Pont{-}Tuset and
                  Federico Perazzi and
                  Sergi Caelles and
                  Pablo Arbel{\'{a}}ez and
                  Alexander Sorkine{-}Hornung and
                  Luc Van Gool},
  title        = {The 2017 {DAVIS} Challenge on Video Object Segmentation},
  journal      = {CoRR},
  volume       = {abs/1704.00675},
  year         = {2017}
}

@inproceedings{DBLP:conf/accv/KhorevaRS18,
  author       = {Anna Khoreva and
                  Anna Rohrbach and
                  Bernt Schiele},
  title        = {Video Object Segmentation with Language Referring Expressions},
  booktitle    = {{ACCV} {(4)}},
  series       = {Lecture Notes in Computer Science},
  volume       = {11364},
  pages        = {123--141},
  publisher    = {Springer},
  year         = {2018}
}

@inproceedings{DBLP:conf/nips/Bai0MWGClZS24,
  author       = {Zechen Bai and
                  Tong He and
                  Haiyang Mei and
                  Pichao Wang and
                  Ziteng Gao and
                  Joya Chen and
                  liulei and
                  Zheng Zhang and
                  Mike Zheng Shou},
  title        = {One Token to Seg Them All: Language Instructed Reasoning Segmentation
                  in Videos},
  booktitle    = {NeurIPS},
  year         = {2024}
}

@article{DBLP:journals/corr/abs-2501-04001,
  author       = {Haobo Yuan and
                  Xiangtai Li and
                  Tao Zhang and
                  Zilong Huang and
                  Shilin Xu and
                  Shunping Ji and
                  Yunhai Tong and
                  Lu Qi and
                  Jiashi Feng and
                  Ming{-}Hsuan Yang},
  title        = {Sa2VA: Marrying {SAM2} with LLaVA for Dense Grounded Understanding
                  of Images and Videos},
  journal      = {CoRR},
  volume       = {abs/2501.04001},
  year         = {2025}
}

@inproceedings{DBLP:conf/cvpr/LaiTCLY0J24,
  author       = {Xin Lai and
                  Zhuotao Tian and
                  Yukang Chen and
                  Yanwei Li and
                  Yuhui Yuan and
                  Shu Liu and
                  Jiaya Jia},
  title        = {{LISA:} Reasoning Segmentation via Large Language Model},
  booktitle    = {{CVPR}},
  pages        = {9579--9589},
  publisher    = {{IEEE}},
  year         = {2024}
}

@inproceedings{DBLP:conf/cvpr/ZhouZPFB017,
  author       = {Bolei Zhou and
                  Hang Zhao and
                  Xavier Puig and
                  Sanja Fidler and
                  Adela Barriuso and
                  Antonio Torralba},
  title        = {Scene Parsing through {ADE20K} Dataset},
  booktitle    = {{CVPR}},
  pages        = {5122--5130},
  publisher    = {{IEEE} Computer Society},
  year         = {2017}
}

@inproceedings{DBLP:conf/cvpr/CaesarUF18,
  author       = {Holger Caesar and
                  Jasper R. R. Uijlings and
                  Vittorio Ferrari},
  title        = {COCO-Stuff: Thing and Stuff Classes in Context},
  booktitle    = {{CVPR}},
  pages        = {1209--1218},
  publisher    = {Computer Vision Foundation / {IEEE} Computer Society},
  year         = {2018}
}

@inproceedings{DBLP:conf/cvpr/RamanathanKPWZG23,
  author       = {Vignesh Ramanathan and
                  Anmol Kalia and
                  Vladan Petrovic and
                  Yi Wen and
                  Baixue Zheng and
                  Baishan Guo and
                  Rui Wang and
                  Aaron Marquez and
                  Rama Kovvuri and
                  Abhishek Kadian and
                  Amir Mousavi and
                  Yiwen Song and
                  Abhimanyu Dubey and
                  Dhruv Mahajan},
  title        = {{PACO:} Parts and Attributes of Common Objects},
  booktitle    = {{CVPR}},
  pages        = {7141--7151},
  publisher    = {{IEEE}},
  year         = {2023}
}

@inproceedings{DBLP:conf/cvpr/ChenMLFUY14,
  author       = {Xianjie Chen and
                  Roozbeh Mottaghi and
                  Xiaobai Liu and
                  Sanja Fidler and
                  Raquel Urtasun and
                  Alan L. Yuille},
  title        = {Detect What You Can: Detecting and Representing Objects Using Holistic
                  Models and Body Parts},
  booktitle    = {{CVPR}},
  pages        = {1979--1986},
  publisher    = {{IEEE} Computer Society},
  year         = {2014}
}

@inproceedings{DBLP:conf/emnlp/KazemzadehOMB14,
  author       = {Sahar Kazemzadeh and
                  Vicente Ordonez and
                  Mark Matten and
                  Tamara L. Berg},
  title        = {ReferItGame: Referring to Objects in Photographs of Natural Scenes},
  booktitle    = {{EMNLP}},
  pages        = {787--798},
  publisher    = {{ACL}},
  year         = {2014}
}

@inproceedings{DBLP:conf/cvpr/MaoHTCY016,
  author       = {Junhua Mao and
                  Jonathan Huang and
                  Alexander Toshev and
                  Oana Camburu and
                  Alan L. Yuille and
                  Kevin Murphy},
  title        = {Generation and Comprehension of Unambiguous Object Descriptions},
  booktitle    = {{CVPR}},
  pages        = {11--20},
  publisher    = {{IEEE} Computer Society},
  year         = {2016}
}

@inproceedings{DBLP:conf/eccv/SeoLH20,
  author       = {Seonguk Seo and
                  Joon{-}Young Lee and
                  Bohyung Han},
  title        = {{URVOS:} Unified Referring Video Object Segmentation Network with
                  a Large-Scale Benchmark},
  booktitle    = {{ECCV} {(15)}},
  series       = {Lecture Notes in Computer Science},
  volume       = {12360},
  pages        = {208--223},
  publisher    = {Springer},
  year         = {2020}
}

@article{rasheedllava++,
  title={LLaVA++: extending visual capabilities with LLaMA-3 and Phi-3 (2024)},
  year={2024},
  author={Rasheed, Hanoona and Maaz, Muhammad and Khan, Salman and Khan, Fahad S},
  journal={URL https://github. com/mbzuai-oryx/LLaVA-pp}
}

@inproceedings{DBLP:conf/iccv/KirillovMRMRGXW23,
  author       = {Alexander Kirillov and
                  Eric Mintun and
                  Nikhila Ravi and
                  Hanzi Mao and
                  Chlo{\'{e}} Rolland and
                  Laura Gustafson and
                  Tete Xiao and
                  Spencer Whitehead and
                  Alexander C. Berg and
                  Wan{-}Yen Lo and
                  Piotr Doll{\'{a}}r and
                  Ross B. Girshick},
  title        = {Segment Anything},
  booktitle    = {{ICCV}},
  pages        = {3992--4003},
  publisher    = {{IEEE}},
  year         = {2023}
}

@misc{ISAT_with_segment_anything,
  title={{ISAT with Segment Anything: An Interactive Semi-Automatic Annotation Tool}},
  author={Ji, Shuwei and Zhang, Hongyuan},
  url={https://github.com/yatengLG/ISAT_with_segment_anything},
  note={Updated on 2025-02-07},
  year={2024},
  version={1.29}
}

@article{DBLP:journals/corr/abs-2408-00714,
  author       = {Nikhila Ravi and
                  Valentin Gabeur and
                  Yuan{-}Ting Hu and
                  Ronghang Hu and
                  Chaitanya Ryali and
                  Tengyu Ma and
                  Haitham Khedr and
                  Roman R{\"{a}}dle and
                  Chlo{\'{e}} Rolland and
                  Laura Gustafson and
                  Eric Mintun and
                  Junting Pan and
                  Kalyan Vasudev Alwala and
                  Nicolas Carion and
                  Chao{-}Yuan Wu and
                  Ross B. Girshick and
                  Piotr Doll{\'{a}}r and
                  Christoph Feichtenhofer},
  title        = {{SAM} 2: Segment Anything in Images and Videos},
  journal      = {CoRR},
  volume       = {abs/2408.00714},
  year         = {2024}
}

@inproceedings{zhang2024llava,
  title={Llava-grounding: Grounded visual chat with large multimodal models},
  author={Zhang, Hao and Li, Hongyang and Li, Feng and Ren, Tianhe and Zou, Xueyan and Liu, Shilong and Huang, Shijia and Gao, Jianfeng and Leizhang and Li, Chunyuan and others},
  booktitle={European Conference on Computer Vision},
  pages={19--35},
  year={2024},
  organization={Springer}
}

@article{ren2024grounded,
  title={Grounded sam: Assembling open-world models for diverse visual tasks},
  author={Ren, Tianhe and Liu, Shilong and Zeng, Ailing and Lin, Jing and Li, Kunchang and Cao, He and Chen, Jiayu and Huang, Xinyu and Chen, Yukang and Yan, Feng and others},
  journal={arXiv preprint arXiv:2401.14159},
  year={2024}
}

@misc{gpt4o,
  author = {OpenAI},
  title = {Hello GPT-4o},
  howpublished = {\url{https://openai.com/index/hello-gpt-4o/}},
year      = {2024},
  note = {Accessed: 2024-07-29}
}

@inproceedings{liu2024grounding,
  title={Grounding dino: Marrying dino with grounded pre-training for open-set object detection},
  author={Liu, Shilong and Zeng, Zhaoyang and Ren, Tianhe and Li, Feng and Zhang, Hao and Yang, Jie and Jiang, Qing and Li, Chunyuan and Yang, Jianwei and Su, Hang and others},
  booktitle={European Conference on Computer Vision},
  pages={38--55},
  year={2024},
  organization={Springer}
}

@article{DBLP:journals/tcsv/TangLLLJJYX22,
  author       = {Zongheng Tang and
                  Yue Liao and
                  Si Liu and
                  Guanbin Li and
                  Xiaojie Jin and
                  Hongxu Jiang and
                  Qian Yu and
                  Dong Xu},
  title        = {Human-Centric Spatio-Temporal Video Grounding With Visual Transformers},
  journal      = {{IEEE} Trans. Circuits Syst. Video Technol.},
  volume       = {32},
  number       = {12},
  pages        = {8238--8249},
  year         = {2022}
}

@inproceedings{DBLP:conf/cvpr/ZhangZZWLG20,
  author       = {Zhu Zhang and
                  Zhou Zhao and
                  Yang Zhao and
                  Qi Wang and
                  Huasheng Liu and
                  Lianli Gao},
  title        = {Where Does It Exist: Spatio-Temporal Video Grounding for Multi-Form
                  Sentences},
  booktitle    = {{CVPR}},
  pages        = {10665--10674},
  publisher    = {Computer Vision Foundation / {IEEE}},
  year         = {2020}
}

@inproceedings{DBLP:conf/cvpr/Gu00L024,
  author       = {Xin Gu and
                  Heng Fan and
                  Yan Huang and
                  Tiejian Luo and
                  Libo Zhang},
  title        = {Context-Guided Spatio-Temporal Video Grounding},
  booktitle    = {{CVPR}},
  pages        = {18330--18339},
  publisher    = {{IEEE}},
  year         = {2024}
}

@inproceedings{DBLP:conf/nips/JinLYM22,
  author       = {Yang Jin and
                  Yongzhi Li and
                  Zehuan Yuan and
                  Yadong Mu},
  title        = {Embracing Consistency: {A} One-Stage Approach for Spatio-Temporal
                  Video Grounding},
  booktitle    = {NeurIPS},
  year         = {2022}
}

@article{DBLP:journals/corr/abs-2107-09609,
  author       = {Jie Lei and
                  Tamara L. Berg and
                  Mohit Bansal},
  title        = {QVHighlights: Detecting Moments and Highlights in Videos via Natural
                  Language Queries},
  journal      = {CoRR},
  volume       = {abs/2107.09609},
  year         = {2021}
}

@inproceedings{DBLP:conf/eccv/YanWYJHKXG24,
  author       = {Cilin Yan and
                  Haochen Wang and
                  Shilin Yan and
                  Xiaolong Jiang and
                  Yao Hu and
                  Guoliang Kang and
                  Weidi Xie and
                  Efstratios Gavves},
  title        = {{VISA:} Reasoning Video Object Segmentation via Large Language Models},
  booktitle    = {{ECCV} {(15)}},
  series       = {Lecture Notes in Computer Science},
  volume       = {15073},
  pages        = {98--115},
  publisher    = {Springer},
  year         = {2024}
}

@inproceedings{DBLP:conf/cvpr/DiX24,
  author       = {Shangzhe Di and
                  Weidi Xie},
  title        = {Grounded Question-Answering in Long Egocentric Videos},
  booktitle    = {{CVPR}},
  pages        = {12934--12943},
  publisher    = {{IEEE}},
  year         = {2024}
}

@article{DBLP:journals/corr/abs-2410-07177,
  author       = {Hanrong Ye and
                  Haotian Zhang and
                  Erik A. Daxberger and
                  Lin Chen and
                  Zongyu Lin and
                  Yanghao Li and
                  Bowen Zhang and
                  Haoxuan You and
                  Dan Xu and
                  Zhe Gan and
                  Jiasen Lu and
                  Yinfei Yang},
  title        = {MM-Ego: Towards Building Egocentric Multimodal LLMs},
  journal      = {CoRR},
  volume       = {abs/2410.07177},
  year         = {2024}
}

@inproceedings{DBLP:conf/nips/ChandrasegaranG24,
  author       = {Keshigeyan Chandrasegaran and
                  Agrim Gupta and
                  Lea M. Hadzic and
                  Taran Kota and
                  Jimming He and
                  Crist{\'{o}}bal Eyzaguirre and
                  Zane Durante and
                  Manling Li and
                  Jiajun Wu and
                  Li Fei{-}Fei},
  title        = {HourVideo: 1-Hour Video-Language Understanding},
  booktitle    = {NeurIPS},
  year         = {2024}
}

@inproceedings{gavrilyuk2018actor,
  title={Actor and action video segmentation from a sentence},
  author={Gavrilyuk, Kirill and Ghodrati, Amir and Li, Zhenyang and Snoek, Cees GM},
  booktitle={Proceedings of the IEEE conference on computer vision and pattern recognition},
  pages={5958--5966},
  year={2018}
}

@article{DBLP:journals/tist/YaoLXZZ20,
  author       = {Rui Yao and
                  Guosheng Lin and
                  Shixiong Xia and
                  Jiaqi Zhao and
                  Yong Zhou},
  title        = {Video Object Segmentation and Tracking: {A} Survey},
  journal      = {{ACM} Trans. Intell. Syst. Technol.},
  volume       = {11},
  number       = {4},
  pages        = {36:1--36:47},
  year         = {2020}
}

@inproceedings{mttr,
  title={End-to-end referring video object segmentation with multimodal transformers},
  author={Botach, Adam and Zheltonozhskii, Evgenii and Baskin, Chaim},
  booktitle={Proceedings of the IEEE/CVF Conference on Computer Vision and Pattern Recognition},
  pages={4985--4995},
  year={2022}
}

@inproceedings{wu2023onlinerefer,
  title={Onlinerefer: A simple online baseline for referring video object segmentation},
  author={Wu, Dongming and Wang, Tiancai and Zhang, Yuang and Zhang, Xiangyu and Shen, Jianbing},
  booktitle={Proceedings of the IEEE/CVF International Conference on Computer Vision},
  pages={2761--2770},
  year={2023}
}

@inproceedings{li2023robust,
  title={Robust referring video object segmentation with cyclic structural consensus},
  author={Li, Xiang and Wang, Jinglu and Xu, Xiaohao and Li, Xiao and Raj, Bhiksha and Lu, Yan},
  booktitle={Proceedings of the IEEE/CVF International Conference on Computer Vision},
  pages={22236--22245},
  year={2023}
}

@inproceedings{wu2022referformer,
  title={Language as queries for referring video object segmentation},
  author={Wu, Jiannan and Jiang, Yi and Sun, Peize and Yuan, Zehuan and Luo, Ping},
  booktitle={Proceedings of the IEEE/CVF Conference on Computer Vision and Pattern Recognition},
  pages={4974--4984},
  year={2022}
}

@inproceedings{miao2023spectrum,
  title={Spectrum-guided multi-granularity referring video object segmentation},
  author={Miao, Bo and Bennamoun, Mohammed and Gao, Yongsheng and Mian, Ajmal},
  booktitle={Proceedings of the IEEE/CVF International Conference on Computer Vision},
  pages={920--930},
  year={2023}
}

@inproceedings{DBLP:conf/cvpr/PerazziPMGGS16,
  author       = {Federico Perazzi and
                  Jordi Pont{-}Tuset and
                  Brian McWilliams and
                  Luc Van Gool and
                  Markus H. Gross and
                  Alexander Sorkine{-}Hornung},
  title        = {A Benchmark Dataset and Evaluation Methodology for Video Object Segmentation},
  booktitle    = {{CVPR}},
  pages        = {724--732},
  publisher    = {{IEEE} Computer Society},
  year         = {2016}
}

@article{DBLP:journals/corr/abs-1809-03327,
  author       = {Ning Xu and
                  Linjie Yang and
                  Yuchen Fan and
                  Dingcheng Yue and
                  Yuchen Liang and
                  Jianchao Yang and
                  Thomas S. Huang},
  title        = {YouTube-VOS: {A} Large-Scale Video Object Segmentation Benchmark},
  journal      = {CoRR},
  volume       = {abs/1809.03327},
  year         = {2018}
}

@inproceedings{egoschema,
  author       = {Karttikeya Mangalam and
                  Raiymbek Akshulakov and
                  Jitendra Malik},
  title        = {EgoSchema: {A} Diagnostic Benchmark for Very Long-form Video Language
                  Understanding},
  booktitle    = {NeurIPS},
  year         = {2023}
}

@inproceedings{ego4dgoalstep,
  author       = {Yale Song and
                  Eugene Byrne and
                  Tushar Nagarajan and
                  Huiyu Wang and
                  Miguel Martin and
                  Lorenzo Torresani},
  title        = {Ego4D Goal-Step: Toward Hierarchical Understanding of Procedural Activities},
  booktitle    = {NeurIPS},
  year         = {2023}
}

@inproceedings{QAEGO4D,
  author       = {Shangzhe Di and
                  Weidi Xie},
  title        = {Grounded Question-Answering in Long Egocentric Videos},
  booktitle    = {{CVPR}},
  pages        = {12934--12943},
  publisher    = {{IEEE}},
  year         = {2024}
}

@article{egovideo,
  author       = {Baoqi Pei and
                  Guo Chen and
                  Jilan Xu and
                  Yuping He and
                  Yicheng Liu and
                  Kanghua Pan and
                  Yifei Huang and
                  Yali Wang and
                  Tong Lu and
                  Limin Wang and
                  Yu Qiao},
  title        = {EgoVideo: Exploring Egocentric Foundation Model and Downstream Adaptation},
  journal      = {CoRR},
  volume       = {abs/2406.18070},
  year         = {2024}
}

@inproceedings{egovlp,
  author       = {Kevin Qinghong Lin and
                  Jinpeng Wang and
                  Mattia Soldan and
                  Michael Wray and
                  Rui Yan and
                  Eric Zhongcong Xu and
                  Difei Gao and
                  Rong{-}Cheng Tu and
                  Wenzhe Zhao and
                  Weijie Kong and
                  Chengfei Cai and
                  Hongfa Wang and
                  Dima Damen and
                  Bernard Ghanem and
                  Wei Liu and
                  Mike Zheng Shou},
  title        = {Egocentric Video-Language Pretraining},
  booktitle    = {NeurIPS},
  year         = {2022}
}

@inproceedings{egovlpv2,
  author       = {Shraman Pramanick and
                  Yale Song and
                  Sayan Nag and
                  Kevin Qinghong Lin and
                  Hardik Shah and
                  Mike Zheng Shou and
                  Rama Chellappa and
                  Pengchuan Zhang},
  title        = {EgoVLPv2: Egocentric Video-Language Pre-training with Fusion in the
                  Backbone},
  booktitle    = {{ICCV}},
  pages        = {5262--5274},
  publisher    = {{IEEE}},
  year         = {2023}
}

@inproceedings{EmbodiedGPT,
  author       = {Yao Mu and
                  Qinglong Zhang and
                  Mengkang Hu and
                  Wenhai Wang and
                  Mingyu Ding and
                  Jun Jin and
                  Bin Wang and
                  Jifeng Dai and
                  Yu Qiao and
                  Ping Luo},
  title        = {EmbodiedGPT: Vision-Language Pre-Training via Embodied Chain of Thought},
  booktitle    = {NeurIPS},
  year         = {2023}
}

@article{DBLP:journals/ijcv/PlizzariGFBRFDT24,
  author       = {Chiara Plizzari and
                  Gabriele Goletto and
                  Antonino Furnari and
                  Siddhant Bansal and
                  Francesco Ragusa and
                  Giovanni Maria Farinella and
                  Dima Damen and
                  Tatiana Tommasi},
  title        = {An Outlook into the Future of Egocentric Vision},
  journal      = {Int. J. Comput. Vis.},
  volume       = {132},
  number       = {11},
  pages        = {4880--4936},
  year         = {2024}
}

@inproceedings{jang2017tgif,
  title={Tgif-qa: Toward spatio-temporal reasoning in visual question answering},
  author={Jang, Yunseok and Song, Yale and Yu, Youngjae and Kim, Youngjin and Kim, Gunhee},
  booktitle={Proceedings of the IEEE conference on computer vision and pattern recognition},
  pages={2758--2766},
  year={2017}
}

@inproceedings{xu2017video,
  title={Video question answering via gradually refined attention over appearance and motion},
  author={Xu, Dejing and Zhao, Zhou and Xiao, Jun and Wu, Fei and Zhang, Hanwang and He, Xiangnan and Zhuang, Yueting},
  booktitle={Proceedings of the 25th ACM international conference on Multimedia},
  pages={1645--1653},
  year={2017}
}

@inproceedings{xiao2021next,
  title={Next-qa: Next phase of question-answering to explaining temporal actions},
  author={Xiao, Junbin and Shang, Xindi and Yao, Angela and Chua, Tat-Seng},
  booktitle={Proceedings of the IEEE/CVF conference on computer vision and pattern recognition},
  pages={9777--9786},
  year={2021}
}

@inproceedings{yu2019activitynet,
  title={Activitynet-qa: A dataset for understanding complex web videos via question answering},
  author={Yu, Zhou and Xu, Dejing and Yu, Jun and Yu, Ting and Zhao, Zhou and Zhuang, Yueting and Tao, Dacheng},
  booktitle={Proceedings of the AAAI Conference on Artificial Intelligence},
  volume={33},
  pages={9127--9134},
  year={2019}
}

@inproceedings{llava15,
  author       = {Haotian Liu and
                  Chunyuan Li and
                  Yuheng Li and
                  Yong Jae Lee},
  title        = {Improved Baselines with Visual Instruction Tuning},
  booktitle    = {{CVPR}},
  pages        = {26286--26296},
  publisher    = {{IEEE}},
  year         = {2024}
}

@article{qwen2vl,
  author       = {Peng Wang and
                  Shuai Bai and
                  Sinan Tan and
                  Shijie Wang and
                  Zhihao Fan and
                  Jinze Bai and
                  Keqin Chen and
                  Xuejing Liu and
                  Jialin Wang and
                  Wenbin Ge and
                  Yang Fan and
                  Kai Dang and
                  Mengfei Du and
                  Xuancheng Ren and
                  Rui Men and
                  Dayiheng Liu and
                  Chang Zhou and
                  Jingren Zhou and
                  Junyang Lin},
  title        = {Qwen2-VL: Enhancing Vision-Language Model's Perception of the
                  World at Any Resolution},
  journal      = {CoRR},
  volume       = {abs/2409.12191},
  year         = {2024}
}

@inproceedings{hudson2019gqa,
  title={Gqa: A new dataset for real-world visual reasoning and compositional question answering},
  author={Hudson, Drew A and Manning, Christopher D},
  booktitle={CVPR},
  year={2019}
}

@article{yue2023mmmu,
  title={Mmmu: A massive multi-discipline multimodal understanding and reasoning benchmark for expert agi},
  author={Yue, Xiang and Ni, Yuansheng and Zhang, Kai and Zheng, Tianyu and Liu, Ruoqi and Zhang, Ge and Stevens, Samuel and Jiang, Dongfu and Ren, Weiming and Sun, Yuxuan and others},
  journal={arXiv preprint arXiv:2311.16502},
  year={2023}
}

@article{yu2023mm,
  title={Mm-vet: Evaluating large multimodal models for integrated capabilities},
  author={Yu, Weihao and Yang, Zhengyuan and Li, Linjie and Wang, Jianfeng and Lin, Kevin and Liu, Zicheng and Wang, Xinchao and Wang, Lijuan},
  journal={arXiv preprint arXiv:2308.02490},
  year={2023}
}

@inproceedings{DBLP:conf/iclr/LoshchilovH19,
  author       = {Ilya Loshchilov and
                  Frank Hutter},
  title        = {Decoupled Weight Decay Regularization},
  booktitle    = {{ICLR} (Poster)},
  publisher    = {OpenReview.net},
  year         = {2019}
}

@inproceedings{DBLP:conf/kdd/RasleyRRH20,
  author       = {Jeff Rasley and
                  Samyam Rajbhandari and
                  Olatunji Ruwase and
                  Yuxiong He},
  title        = {DeepSpeed: System Optimizations Enable Training Deep Learning Models
                  with Over 100 Billion Parameters},
  booktitle    = {{KDD}},
  pages        = {3505--3506},
  publisher    = {{ACM}},
  year         = {2020}
}

@article{liu2023grounding,
  title={Grounding dino: Marrying dino with grounded pre-training for open-set object detection},
  author={Liu, Shilong and Zeng, Zhaoyang and Ren, Tianhe and Li, Feng and Zhang, Hao and Yang, Jie and Li, Chunyuan and Yang, Jianwei and Su, Hang and Zhu, Jun and others},
  journal={arXiv preprint arXiv:2303.05499},
  year={2023}
}

@inproceedings{VISOR2022,
           title={EPIC-KITCHENS VISOR Benchmark: VIdeo Segmentations and Object Relations},
           author={Darkhalil, Ahmad and Shan, Dandan and Zhu, Bin and Ma, Jian and Kar, Amlan and Higgins, Richard and Fidler, Sanja and Fouhey, David and Damen, Dima},
           booktitle   = {Proceedings of the Neural Information Processing Systems (NeurIPS) Track on Datasets and Benchmarks},
           year      = {2022}
}

@InProceedings{perrett2025hdepic,
  author    = {Perrett, Toby and Darkhalil, Ahmad and Sinha, Saptarshi and Emara, Omar and Pollard, Sam and Parida, Kranti and Liu, Kaiting and Gatti, Prajwal and Bansal, Siddhant and Flanagan, Kevin and Chalk, Jacob and Zhu, Zhifan and Guerrier, Rhodri and Abdelazim, Fahd and Zhu, Bin and Moltisanti, Davide and Wray, Michael and Doughty, Hazel and Damen, Dima},
  title     = {HD-EPIC: A Highly-Detailed Egocentric Video Dataset},
  booktitle = {Proceedings of the IEEE/CVF Conference on Computer Vision and Pattern Recognition (CVPR)},
  year      = {2025},
  month     = {June}
}

@inproceedings{zhong-2024-beyond,
    title = "Beyond Embeddings: The Promise of Visual Table in Visual Reasoning",
    author = "Zhong, Yiwu  and
      Hu, Zi-Yuan  and
      Lyu, Michael  and
      Wang, Liwei",
    booktitle = "Proceedings of the 2024 Conference on Empirical Methods in Natural Language Processing",
    month = nov,
    year = "2024",
    publisher = "Association for Computational Linguistics",
    pages = "6876--6911",
}
}

\newpage
\appendix

\smallskip
\noindent {\LARGE {{\rm {\bf Appendix}}}}

\bigskip

In the appendix, we provide more details in addition to our main paper: (1) comparison of existing datasets related to spatiotemporal grounding tasks, 
% (2) more related work discussions, 
(2) verification results of EgoMask annotations, 
(3) additional statistics of our datasets, 
(4) additional experimental results on our benchmark, 
(5) additional details in our annotation pipeline, 
(6) additional analysis on the characteristics of egocentric videos, 
and (7) additional visualization examples.

\begin{table*}[]
\centering
\resizebox{\textwidth}{!}{%
\begin{tabular}{c|c|c|c|c|ccc|c|c}
\toprule
\textbf{Dataset} &
  \textbf{Egocentric} &
  \multicolumn{1}{c|}{\textbf{Video Length (s)}} &
  \multicolumn{1}{c|}{\textbf{Total Duration (\%)}} &
  \multicolumn{1}{c|}{\textbf{BBox Area (\%)}} &
  \multicolumn{1}{c}{\textbf{\# Traj.}} &
  \multicolumn{1}{c}{\textbf{Avg. Traj Length. (\%)}} &
  \multicolumn{1}{c|}{\textbf{Disappear. Ratio (\%)}} &
  \multicolumn{1}{c|}{\textbf{Adj. Bbox IoU (\%)}} &
  \textbf{Anno. type} \\ 
\midrule
{Egotracks~\cite{DBLP:conf/nips/TangLGF023}}  &  \cmark  & 369.00 & 25.23 & 2.42  & 22.96 & 1.35  & 496.31 & 45.07   & Bbox \\ 
\midrule
{RefEgo~\cite{DBLP:conf/iccv/KuritaKO23}}      &  \cmark  & 12.27  & 76.82 & 2.84  & 2.01  & 50.52 & 24.20  & 22.16 & Bbox \\ 
\midrule
{Mevis~\cite{DBLP:conf/iccv/Ding0H0L23}}      &  \xmark  & 69.83  & 77.78 & 10.72 & 1.42  & 68.88 & 19.93  & 65.69 & Mask         \\ 
\midrule
{Ref-Davis~\cite{DBLP:conf/accv/KhorevaRS18}}  & \xmark & 69.41  & 95.66 & 13.00 & 1.18  & 89.84 & 4.41   & 83.30 & Mask         \\ 
\midrule
{Ref-YT-VOS~\cite{DBLP:conf/eccv/SeoLH20}} & \xmark & 26.53  & 93.57 & 18.49 & 1.11  & 89.65 & 5.97   & 72.81 & Mask         \\ 

% 369.00 & 25.23 & 2.42  & 22.96 & 1.35  & 496.31 & 45.07 \\
% 12.27  & 76.82 & 2.84  & 2.01  & 50.52 & 24.20  & 22.16 \\
% 69.83  & 77.78 & 10.72 & 1.42  & 68.88 & 19.93  & 65.69 \\
% 69.41  & 95.66 & 13.00 & 1.18  & 89.84 & 4.41   & 83.30 \\
% 26.53  & 93.57 & 18.49 & 1.11  & 89.65 & 5.97   & 72.81

\bottomrule
\end{tabular}
}
\caption{Comparison of existing datasets related to spatiotemporal grounding task. 
The ``\textbf{Total Duration}~(\%)'' means the percent of the total appearance of the referred objects. 
The ``BBox Area~(\%)'' means the average area of the annotated bounding box over the frame size, which can reveal the \textbf{object size}.
The ``\# Traj.'' means the number of object's continuous trajectories throughout the video.
The  ``Avg. Traj. Length (\%)'' means the average of each trajectory duration over the whole video and
the ``Disappear. Ratio(\%)'' is formulated as the mean of each disappearance duration over each trajectory duration.
These two values can reveal \textbf{the sparsity of the continuous trajectory}.
the ``Adj. Bbox IoU(\%)'' shows the \textbf{positional shifts} over the adjacent frames by calculating the IoU value of the bounding boxes of the target object.}
\label{table:dataset-comparison}
\end{table*}
\begin{table}[]
\centering
\resizebox{0.90\columnwidth}{!}{%
\begin{tabular}{c|c|c}
\toprule
  \multicolumn{1}{c|}{\textbf{Domains}} &
  \multicolumn{1}{c|}{\textbf{Grounded-SAM2}} &
  \multicolumn{1}{c}{\textbf{Sa2VA-26B}} \\ 
\midrule
Kitchen &  40.19 & 31.07  \\ 
\midrule
Non-Kitchen & 38.59 & 29.95 \\ 

\bottomrule
\end{tabular}
}
\caption{Comparison of model performances on Kitchen and Non-Kitchen domains on EgoMask.}
\label{table:kitchen-comparison}
\end{table}
\section{Comparison of Existing Datasets}
\label{sec:Existing_dataset_comparison}
We present the detailed comparison of some existing datasets related to spatiotemporal grounding tasks in Table~\ref{table:dataset-comparison} to show the distinguished difference between egocentric videos and exocentric videos.

\section{Annotation verification of EgoMask}

We verified 20\% test annotations with three experts to score 1 to 5, where 5 is the best. The average scores of expressions/masks are 4.65/4.92. If we set 3 as the threshold score, the error rate is 2.5\%/0\%, suggesting high quality of our annotations.

\begin{figure}[h]
\begin{center}
\includegraphics[width=.7\linewidth]{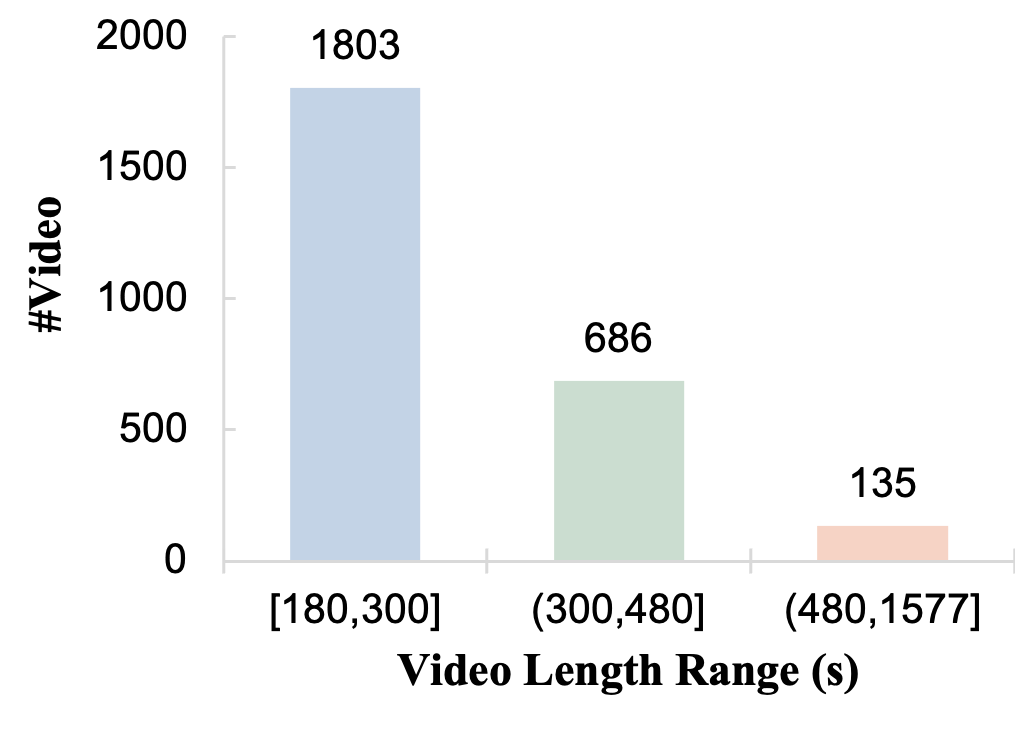}
\end{center}
\vspace{-10pt}
   \caption{Video Length Distribution of \textbf{\trainingdataset{}}.
   } 
   % \vspace{-5pt}
\label{fig:dis_traindata}
\end{figure}

\begin{figure}[h]
\begin{center}
\includegraphics[width=.9\linewidth]{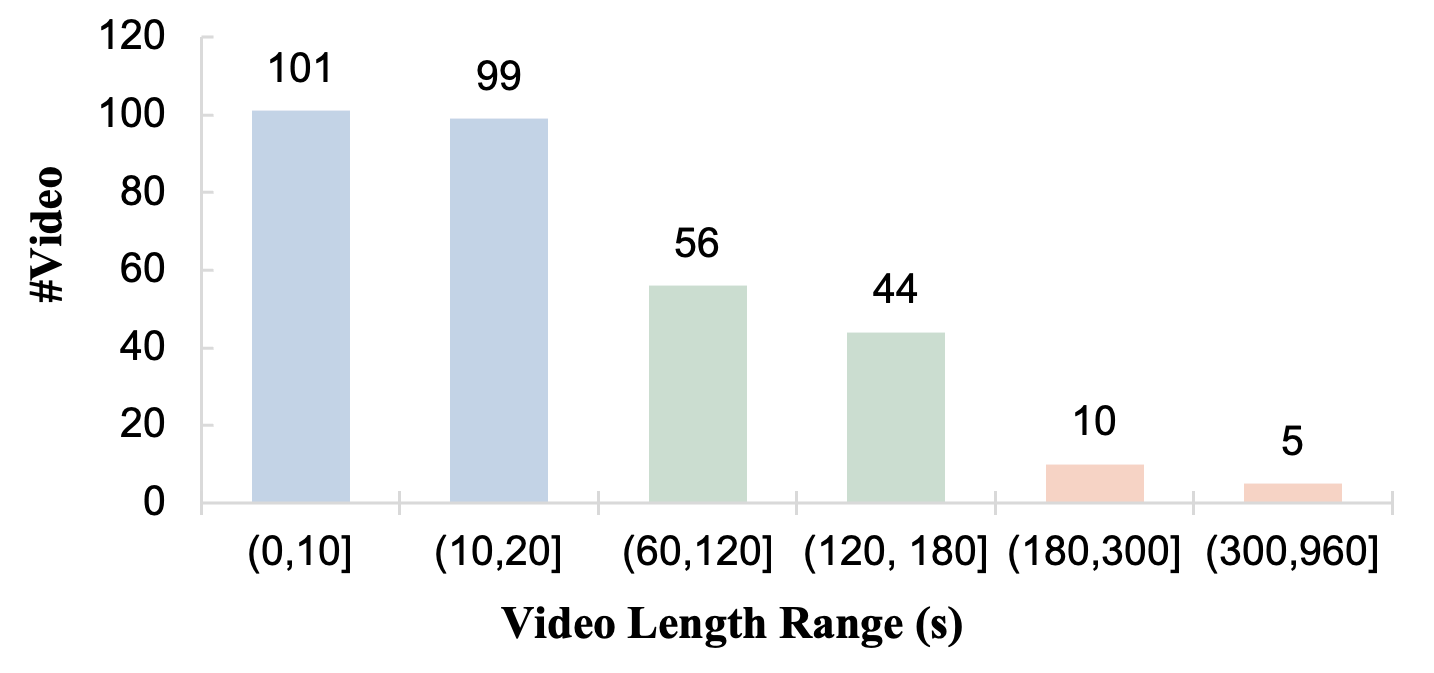}
\end{center}
\vspace{-10pt}
   \caption{Video Length Distribution of \textbf{\benchmark{}}.
   } 
   % \vspace{-5pt}
\label{fig:dis_benchmark}
\end{figure}

\section{Video Length Distributions of our datasets}
% \begin{figure}[]
%   \centering
%   % \vspace{-10pt}
%   \includegraphics[width=0.99\linewidth]{ICCV2025-Arxiv/images/small_length_dis.pdf}
%   \vspace{-10pt}
%    \caption{Video Length Distribution of our test set and train set.}
%    \label{fig:video_dis}
%    \vspace{-15pt}
% \end{figure}
The video length distribution of our training set \textit{\trainingdataset{}} and test set \textit{\benchmark{}} is shown in Figure~\ref{fig:dis_traindata}  and ~\ref{fig:dis_benchmark}. 

\section{Prompts for Expression Generation}
\label{sec:appendix_prompts}

\begin{figure*}[]
\begin{center}
\begin{tcolorbox}[colback=black!5!white, colframe=black!75!black, title= Short and long expression generation, width=0.95\linewidth]
% \vspace{-5pt}
\textcolor{blue}{\{frame\_0\} \{frame\_1\} ...} \\
Please help me generate referring expressions for object segmentation. \\
There are \textcolor{blue}{\{total\_frames\}} frames from a video.
Each frame contains a red bounding box that corresponds to the same object. Based on the object in the red bounding box and its object tag, please generate the descriptions that uniquely identify the object throughout the video.\\
Object tag: \textcolor{blue}{\{object\_category\}} \\
- Output should consist of two lines, separated by a newline: \\
1. A short expression with no more than 10 words, starting with "Short expressions: ".\\
2. A longer expression with more detailed illustrations, starting with "Long expressions: ". \\
\textbf{Restriction Policies}: \\
- The referring expressions should be concise and informative. They can be spatial location in the physical world, OCR characters on the object, spatial relations to surrounding objects, action relations to surrounding objects, relative size compared to surrounding objects, color, geometry shape, material, texture pattern, motion or dynamics of objects, and so on.\\
- The generated referring expressions should clearly identify the object to avoid any ambiguity without referencing bounding boxes in the video. \\
- Do not use "red bounding box", "image", or "frame" in the answer.

% \vspace{-5pt}
\end{tcolorbox}
% \vspace{-10pt}
\caption{Expression generation prompts.}
% \vspace{-15pt}
\label{fig:expression_prompts}
\end{center}
\end{figure*}

\begin{figure*}[]\small
\begin{center}
\begin{tcolorbox}[colback=black!5!white, colframe=black!75!black, title= Object metadata generation, width=0.95\linewidth]
% \vspace{-5pt}
\textcolor{blue}{\{frame\_0\} \{frame\_1\} ...} \\
Please help me generate object descriptions.
These are \textcolor{blue}{\{total\_frames\}} frames from a video. Each frame contains a red bounding box that corresponds to the same object. Based on the object in the red bounding box and its object tag, please generate its caption, visual attributes and affordance description (if applicable). \\
Object tag: \textcolor{blue}{\{object\_category\}} \\
- Output should consist of three lines, separated by a newline:\\
1. A clear object caption with no more than 10 words, starting with "Object Caption: ".\\
2. The visual attributes of the object, starting with "Visual Attributes: ".\\
3. A concrete affordance description of the object, starting with "Object: Affordance: ".\\
\textbf{Restriction Policies}: \\
- Use the provided object tag selectively, as it may contain noise. \\
- The object caption should be a noun phrase.\\
- The object caption should clearly identify the object with minimal words to avoid any ambiguity without referencing bounding boxes. \\
- Visual attributes characterize the objects in images. They can be spatial location in the physical world, OCR characters on the object, spatial relations to surrounding objects, action relations to surrounding objects, relative size compared to surrounding objects, color, geometry shape, material, texture pattern, motion or dynamics of objects, and so on.\\
- The affordance description should focus on the object's potential actions, interactions, or functions, describing how the object can be utilized or manipulated in a given context. Avoid generic statements and provide specific and practical insights into the object's affordances.\\
- The affordance description should be a verb phrase, e.g., cut vegetables, clean the tables, etc. If there is no affordance about the object, output "None". \\
- Do not use "red bounding box", "image", or "frame" in the answer.\\

% \vspace{-5pt}
\end{tcolorbox}

% \vspace{-5pt}
\caption{Prompts for generating metadata of the labeled object.}
% \vspace{-8pt}
\label{fig:object_meta_prompts}
\end{center}
\end{figure*}
% \usepackage{graphicx}
\begin{table}[]
\centering
\resizebox{0.50\columnwidth}{!}{%
\begin{tabular}{cc}
\toprule
Type                                 & Average Length \\
\midrule
\multicolumn{2}{c}{\textit{\textbf{Expression}}} \\ 
\midrule
Short expression                    & 7.75        \\
Long expression                     & 26.31      \\
\midrule

\multicolumn{2}{c}{\textit{\textbf{Metadata}}} \\ 
\midrule

Caption                              & 2.98        \\
Visual attributes                   & 16.19       \\
Affordance                           & 4.72        \\

\bottomrule
\end{tabular}%
}
\caption{Statistics of the generated expressions and metadata.}
\label{table:exp_stat}
\end{table}

To guarantee diversity, we use two different strategies to generate the referring expressions as the language queries for our dataset.
We use the prompt shown in Figure~\ref{fig:expression_prompts} to directly instruct the GPT-4o to generate a short expression and a longer expression. 
We use the prompt shown in Figure~\ref{fig:object_meta_prompts} to first instruct the GPT-4o to generate the metadata of the target objects and then use templates to form expressions.
The length statistics are shown in Figure~\ref{table:exp_stat}.

% \section{Effectiveness of our data annotation pipeline}
% Our test set is fully verified by human annotators, including expressions and masks. 
% Among them, only 8\% generated expressions have major hallucination, 
% This shows the effectiveness of our automatic annotation pipeline.

% Although our training data inevitably contain noises,our experiments validate that models can still learn knowledge from these data. 

\section{Evaluation results of closed-source models}
We conduct experiments with GPT-4o and Gemini-Pro on our benchmark (10\% subset). Due to their inability to support dense segmentation, we prompt them to generate the corners of boxes, which are then evaluated using $\IouGoldPred{}$. The result is 3.47\%/1.47\% for GPT-4o/Gemini, demonstrating their limitations on our task.

\section{Effects of Characteristics of Egocentric Entities}
% \begin{table}[]\small
% \centering
% \resizebox{\columnwidth}{!}{%
% \begin{tabular}{c|c|c|c|cc|cc}
% \toprule
% \multirow{2}{*}{\textbf{Type}} & 
% \multirow{2}{*}{\textbf{Avg.}}  & 
% \textbf{Below} &
% \textbf{\#Test}& 
% \textbf{Grounded} & \textbf{Sa2VA} & \textbf{VideoLISA} & \textbf{VideoLISA} \\

%     &    & \textbf{Avg.} & \textbf{sample}
%     &    \textbf{-SAM2} & \textbf{-4B} &\textbf{-3.8B} & \textbf{-3.8B (+FT) }\\

% \midrule

% \multirow{2}{*}{Short} & \multirow{2}{*}{80.31} &\cmark & 190 & 39.56 & 21.41 & 9.69  & 14.16 \textbf{(+4.48)}       \\
%                        & & \xmark & 210 & 58.65 & 35.96 & 25.25 & 31.80  \textbf{(+6.56)}      \\
% \midrule
% \multirow{2}{*}{Medium} & \multirow{2}{*}{37.91}  & \cmark & 116 & 14.87 & 9.94  & 0.80  & 1.28 \textbf{(+0.48)} \\
%                        && \xmark & 84  & 40.76 & 24.92 & 14.51 & 21.94  \textbf{(+7.43) }      \\
% \midrule
% \multirow{2}{*}{Long}  & \multirow{2}{*}{27.48} & \cmark & 62  & 12.95 & 1.14  & 0.54  & 0.48 \textbf{(-0.06)}\\
%                        && \xmark & 38  & 43.74 & 19.86 & 12.56 & 18.00   \textbf{(+5.44)}      \\

% \bottomrule
% \end{tabular}
% }
% \caption{Performance Comparison over different subsets of total durations. The Avg. means the average of total duration (\%).}
% \label{table:effects_of_duration}

% \end{table}

\begin{table*}[]\small
\centering
\resizebox{0.99\textwidth}{!}{%
\begin{tabular}{c|c|c|c|c|ccc|cc}
\toprule
\textbf{Type} & 
% \purp{\textbf{Avg. Total Duration(\%)}}  & 
{\textbf{Avg. Total Duration(\%)}}  & 
\textbf{Below Avg.} &
\textbf{\#Test Sample} & 
\textbf{Grounded-SAM2} & 
\textbf{Sa2VA-26B} & 
\textbf{Sa2VA-4B} & 
\textbf{Sa2VA-4B (+FT)} & 
\textbf{VideoLISA} & 
\textbf{VideoLISA-3.8 (+FT)} \\

\midrule

\multirow{2}{*}{Short} & \multirow{2}{*}{80.31} &\cmark & 190 & 40.14 & 29.15 & 21.47 & 22.10 (+\textbf{0.63}) & 9.71  & 14.13  (+\textbf{4.42})     \\
                       & & \xmark & 210 & 58.84 & 44.67 & 35.81 & 39.00 (+\textbf{3.19}) & 25.21 & 31.71  (+\textbf{6.49})   \\
\midrule
\multirow{2}{*}{Medium} & \multirow{2}{*}{36.69}  & \cmark & 118 & 14.71 & 14.18 & 10.51 & 12.95 (+\textbf{2.44})& 0.85  & 1.29 (+\textbf{0.44})\\
                       && \xmark & 82  & 41.59 & 42.58 & 26.40 & 26.54 (+\textbf{0.15}) & 14.57 & 22.48 (+\textbf{7.91})     \\
\midrule
\multirow{2}{*}{Long}  & \multirow{2}{*}{27.48} & \cmark & 62  & 13.13 & 5.25  & 1.28  & 2.60  (+\textbf{1.32}) & 0.54  & 0.48 (-\textbf{0.06})\\
                       && \xmark & 38  & 43.86 & 25.53 & 19.25 & 17.43 (-\textbf{1.83}) & 12.68 & 18.07 (+\textbf{5.39})     \\

\bottomrule
\end{tabular}
}
% \vspace{-5pt}
\caption{Performance Comparison over different subsets of total durations. The {Avg. Total Duration} means the average of total duration (\%).
% The \purp{Avg. Total Duration} means the average of total duration (\%).
}
\label{table:effects_of_duration}
% \vspace{-8pt}

\end{table*}
% \begin{table}[]\small
% \centering
% \resizebox{\columnwidth}{!}{%
% \begin{tabular}{c|c|c|c|cc|cc}
% \toprule
% \multirow{2}{*}{\textbf{Type}} & 
% \multirow{2}{*}{\textbf{Avg.}}  & 
% \textbf{Below} &
% \textbf{\#Test}& 
% \textbf{Grounded} & \textbf{Sa2VA} & \textbf{VideoLISA} & \textbf{VideoLISA} \\

%     &    & \textbf{Avg.} & \textbf{sample}
%     &    \textbf{-SAM2} & \textbf{-4B} &\textbf{-3.8B} & \textbf{-3.8B (+FT)} \\
\begin{table*}[]\small
\centering
\resizebox{0.99\textwidth}{!}{%
\begin{tabular}{c|c|c|c|c|ccc|cc}
\toprule
\textbf{Type} & 
% \org{\textbf{Avg. Mask Area~(\%)}} & 
{\textbf{Avg. Mask Area~(\%)}} & 
\textbf{Below Avg.} &
\textbf{\#Test Sample} & 
\textbf{Grounded-SAM2} & 
\textbf{Sa2VA-26B} & 
\textbf{Sa2VA-4B} & 
\textbf{Sa2VA-4B (+FT)} & 
\textbf{VideoLISA} & 
\textbf{VideoLISA-3.8 (+FT)} \\
\midrule

\multirow{2}{*}{Short} & \multirow{2}{*}{1.83} &\cmark & 308 & 48.63 & 31.40 & 22.29 & 24.31 (+\textbf{2.02}) & 12.91 & 18.45  (+\textbf{5.54})   \\
                       & & \xmark & 92  & 54.38 & 57.06 & 51.44 & 53.28 (+\textbf{1.83}) & 34.40 & 39.80 (+\textbf{5.39})   \\
\midrule
\multirow{2}{*}{Medium} & \multirow{2}{*}{1.87}  & \cmark & 142 & 17.08 & 20.01 & 12.18 & 16.04 (+\textbf{3.86}) & 5.24  & 8.46 (+\textbf{3.22})\\
                       && \xmark & 58  & 46.91 & 40.07 & 28.88 & 24.59 (-\textbf{4.29}) & 9.50  & 13.69 (+\textbf{4.19})    \\
\midrule
\multirow{2}{*}{Long}  & \multirow{2}{*}{1.86} & \cmark & 76  & 17.18 & 11.10 & 5.27  & 6.97 (+\textbf{1.71})  & 3.03  & 4.67 (+\textbf{1.64}) \\
                       && \xmark & 24  & 48.95 & 18.83 & 17.11 & 12.23 (-\textbf{4.88}) & 11.87 & 15.05  (+\textbf{3.18})    \\

\bottomrule
\end{tabular}
}
\caption{Performance Comparison over different subsets of object size. The {Avg. Mask Area} refers to the average mask area (\%) of the queried objects.
% The \org{Avg. Mask Area} refers to the average mask area (\%) of the queried objects.
}
% \vspace{-5pt}
\label{table:effects_of_size}

\end{table*}
% \begin{table}[]\small
% \centering
% \resizebox{\columnwidth}{!}{%
% \begin{tabular}{c|c|c|c|cc|cc}
% \toprule
% \multirow{2}{*}{\textbf{Type}} & 
% \multirow{2}{*}{\textbf{Avg.}}  & 
% \textbf{Below} &
% \textbf{\#Test}& 
% \textbf{Grounded} & \textbf{Sa2VA} & \textbf{VideoLISA} & \textbf{VideoLISA} \\

%     &    & \textbf{Avg.} & \textbf{sample}
%     &    \textbf{-SAM2} & \textbf{-4B} &\textbf{-3.8B} & \textbf{-3.8B (+FT)} \\
\begin{table*}[]\small
\centering
\resizebox{0.99\textwidth}{!}{%
\begin{tabular}{c|c|c|c|c|ccc|cc}
\toprule
\textbf{Type} & 
% \blue{\textbf{Avg. Traj. Length~(\%)}} &
{\textbf{Avg. Traj. Length~(\%)}} &
\textbf{Below Avg.} &
\textbf{\#Test Sample} & 
\textbf{Grounded-SAM2} & 
\textbf{Sa2VA-26B} & 
\textbf{Sa2VA-4B} & 
\textbf{Sa2VA-4B (+FT)} & 
\textbf{VideoLISA} & 
\textbf{VideoLISA-3.8 (+FT)} \\
\midrule

\multirow{2}{*}{Short} & \multirow{2}{*}{57.13} &\cmark & 196 & 44.17 & 29.94 & 21.72 & 23.65 (+\textbf{1.93}) & 10.56 & 15.54   (+\textbf{4.99})    \\
                       & & \xmark & 204 & 55.52 & 44.37 & 35.99 & 38.01 (+\textbf{2.02}) & 24.86 & 30.86 (+\textbf{6.01})     \\
\midrule
\multirow{2}{*}{Medium} & \multirow{2}{*}{11.52}  & \cmark & 156 & 20.83 & 21.81 & 14.69 & 16.20 (+\textbf{1.51}) & 4.03  & 5.60 (+\textbf{1.57}) \\
                       && \xmark & 44  & 43.08 & 40.08 & 25.30 & 26.76(+\textbf{1.46})  & 15.14 & 25.50 (+\textbf{10.36})     \\
\midrule
\multirow{2}{*}{Long}  & \multirow{2}{*}{1.81} & \cmark & 64  & 18.01 & 4.74  & 0.82  & 2.16 (+\textbf{1.34}) & 1.16  & 1.49 (+\textbf{0.33})\\
                       && \xmark & 36  & 36.89 & 27.56 & 21.07 & 19.03 (-\textbf{2.04}) & 12.26 & 17.25   (+\textbf{5.00})  \\

\bottomrule
\end{tabular}
}
\caption{Performance Comparison over different subsets of our test data over object continuous trajectories. 
The {Avg. Traj. Length} refers to the average of each consecutive appearance duration (\%).
% The \blue{Avg. Traj. Length} refers to the average of each consecutive appearance duration (\%).
}
% \vspace{-15pt}
\label{table:effects_of_traj}

\end{table*}
% \begin{table}[]\small
% \centering
% \resizebox{\columnwidth}{!}{%
% \begin{tabular}{c|c|c|c|cc|cc}
% \toprule
% \multirow{2}{*}{\textbf{Type}} & 
% \multirow{2}{*}{\textbf{Avg.}}  & 
% \textbf{Below} &
% \textbf{\#Test}& 
% \textbf{Grounded} & \textbf{Sa2VA} & \textbf{VideoLISA} & \textbf{VideoLISA} \\

%     &    & \textbf{Avg.} & \textbf{sample}
%     &    \textbf{-SAM2} & \textbf{-4B} &\textbf{-3.8B} & \textbf{-3.8B (+FT)} \\

\begin{table*}[]\small
\centering
\resizebox{0.99\textwidth}{!}{%
\begin{tabular}{c|c|c|c|c|ccc|cc}
\toprule
\textbf{Type} & 
% \green{\textbf{Avg. Disappear. Ratio~(\%)}}  & 
{\textbf{Avg. Disappear. Ratio~(\%)}}  & 
\textbf{Below Avg.} &
\textbf{\#Test Sample} & 
\textbf{Grounded-SAM2} &
\textbf{Sa2VA-26B} & 
\textbf{Sa2VA-4B} & 
\textbf{Sa2VA-4B (+FT)} & 
\textbf{VideoLISA} & 
\textbf{VideoLISA-3.8(+FT)} \\

\midrule

\multirow{2}{*}{Short} & \multirow{2}{*}{21.92} &\cmark & 190 & 58.26 & 45.03 & 37.37 & 40.51 (+\textbf{3.14}) & 25.98 & 32.00  (+\textbf{6.02})    \\
                       & & \xmark & 210 & 42.44 & 30.30 & 21.42 & 22.34 (+\textbf{0.92}) & 10.49 & 15.53 (+\textbf{5.04})    \\
\midrule
\multirow{2}{*}{Medium} & \multirow{2}{*}{179.47}  & \cmark & 88  & 38.76 & 39.97 & 24.69 & 24.83 (+\textbf{0.15}) & 13.69 & 21.11(+\textbf{7.42})  \\
                       && \xmark & 112 & 15.49 & 14.72 & 11.01 & 13.56 (+\textbf{2.56})& 0.81  & 1.23 (+\textbf{0.42})  \\
\midrule
\multirow{2}{*}{Long}  & \multirow{2}{*}{450.29} & \cmark & 46  & 38.43 & 21.09 & 15.91 & 14.40 (-\textbf{1.51}) & 10.66 & 15.05 (+\textbf{4.39})\\
                       && \xmark & 54  & 13.19 & 6.03  & 1.47  & 2.99 (+\textbf{1.52}) & 0.46  & 0.45 (-\textbf{0.01}) \\

\bottomrule
\end{tabular}
}
\vspace{-8pt}
\caption{Performance Comparison over different subsets over the ratio of disappearance over appearance. 
The {Avg. Disappear. Ratio} refers to the mean value of the ratio of average disappearance duration over the average trajectory length.
% The \green{Avg. Disappear. Ratio} refers to the mean value of the ratio of average disappearance duration over the average trajectory length.
}
\vspace{-5pt}
\label{table:effects_of_disappear_ratio}

\end{table*}
% \begin{table}[]\small
% \centering
% \resizebox{\columnwidth}{!}{%
% \begin{tabular}{c|c|c|c|cc|cc}
% \toprule
% \multirow{2}{*}{\textbf{Type}} & 
% \multirow{2}{*}{\textbf{Avg.}}  & 
% \textbf{Below} &
% \textbf{\#Test}& 
% \textbf{Grounded} & \textbf{Sa2VA} & \textbf{VideoLISA} & \textbf{VideoLISA} \\

%     &    & \textbf{Avg.} & \textbf{sample}
%     &    \textbf{-SAM2} & \textbf{-4B} &\textbf{-3.8B} & \textbf{-3.8B (+FT)} \\
\begin{table*}[]\small
\centering
\resizebox{0.99\textwidth}{!}{%
\begin{tabular}{c|c|c|c|c|ccc|cc}
\toprule
\textbf{Type} & 
{\textbf{Avg. Adj. Mask IoU (\%)}}  & 
% \teal{\textbf{Avg. Adj. Mask IoU (\%)}}  & 
\textbf{Below Avg.} &
\textbf{\#Test Sample} & 
\textbf{Grounded-SAM2} & 
\textbf{Sa2VA-26B} & 
\textbf{Sa2VA-4B} & 
\textbf{Sa2VA-4B (+FT)} & 
\textbf{VideoLISA} & 
\textbf{VideoLISA-3.8(+FT)} \\

\midrule

\multirow{2}{*}{Short} & \multirow{2}{*}{8.51} &\cmark & 268 & 45.78 & 28.91 & 19.83 & 21.88 (+\textbf{2.05}) & 13.29 & 18.75 (+\textbf{5.46}) \\
                       & & \xmark & 132 & 58.43 & 54.34 & 47.62 & 49.43 (+\textbf{1.81}) & 27.12 & 32.71 (+\textbf{5.59}) \\
\midrule
\multirow{2}{*}{Medium} & \multirow{2}{*}{20.98}  & \cmark & 116 & 16.70 & 19.00 & 12.99 & 14.93 (+\textbf{1.94}) & 1.54  & 3.09 (+\textbf{1.54}) \\
                       && \xmark & 84  & 38.20 & 35.26 & 22.59 & 23.48 (+\textbf{0.89}) & 13.29 & 19.50 (+\textbf{6.21})  \\
\midrule
\multirow{2}{*}{Long}  & \multirow{2}{*}{19.53} & \cmark & 58  & 18.31 & 7.24  & 1.65  & 3.31 (+\textbf{1.66}) & 2.23  & 2.75 (+\textbf{0.52})\\
                       && \xmark & 42  & 33.77 & 20.85 & 17.03 & 15.04 (-\textbf{1.99}) & 9.19  & 13.26 (+\textbf{4.07})  \\

\bottomrule
\end{tabular}
}
\vspace{-8pt}
\caption{Performance Comparison over different subsets over object position shifts. 
The {Avg. Adj. Mask IoU} refers to the mean IoU value of spatial position over the adjacent frames.}
% The \teal{Avg. Adj. Mask IoU} refers to the mean IoU value of spatial position over the adjacent frames.}
\vspace{-5pt}
\label{table:effects_of_position_shifts}

\end{table*}
We further provide an in-depth analysis of four key factors. Specifically, we investigate the relations between model performance and the key factors in each benchmark subset.

\smallskip 
\noindent \textbf{Total duration.}
Table~\ref{table:effects_of_duration} shows the effects of total duration. It is defined as the ratio of total appearance time over the whole video. For all types of benchmarks, the model performs better when the objects have larger total durations (see Below Avg. \xmark).

\smallskip 
\noindent \textbf{Object size.}
Table~\ref{table:effects_of_size} shows the effects of object size. Generally, the model achieves higher performance when the referred objects have larger sizes (see Below Avg. \xmark).

\smallskip 
\noindent \textbf{Continuous trajectories.}
Table~\ref{table:effects_of_traj} and Table~\ref{table:effects_of_disappear_ratio} show the effects of continuous trajectories. The trajectory is defined as one consecutive appearance, and the trajectory length is calculated as the average time of each appearance over the whole video. 
We define the non-trajectory length as the average time of each disappearance over the whole video. 
And then, the ratio of disappearance over appearance is calculated as the ratio of non-trajectory length over trajectory length. 
When the average trajectory length is longer (see Below Avg. \xmark ~ in Table~\ref{table:effects_of_traj}) and with less disappearance (see Below Avg. \cmark ~ in Table~\ref{table:effects_of_disappear_ratio}), the model performs better.

\smallskip 
\noindent \textbf{Positional shift.}
Table~\ref{table:effects_of_position_shifts} shows the effects of positional shifts. When the objects have fewer shifts in the video (see Below Avg. \xmark ~ in Table~\ref{table:effects_of_position_shifts}), the performance improves a lot. 
 
Based on the above analysis, we can safely deduce that spatiotemporal grounding on egocentric videos is much harder than that in exocentric videos. 
We also notice that in most cases, our fine-tuned models, Sa2VA-4B(+FT) and VideoLISA-3.8B(+FT), surpass their pre-trained models. It can verify the effectiveness of our proposed training dataset \trainingdataset{}.

\begin{figure}[ht]
\begin{center}
\includegraphics[width=0.95\linewidth]{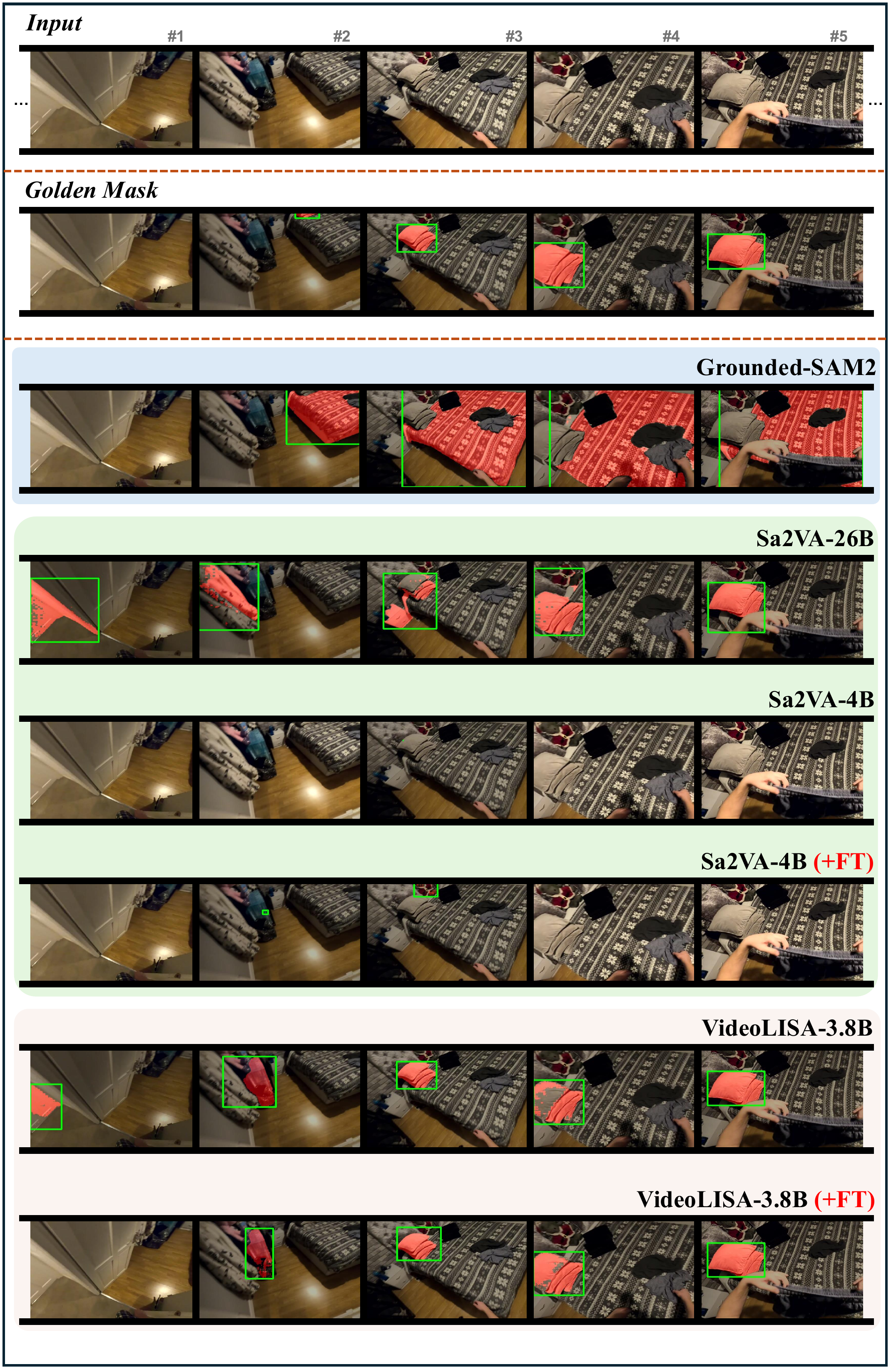}
\end{center}
\vspace{-10pt}
   \caption{Visualization of example from \benchmark{}-Short. The language query is ``the pillows stacked on top of bed''.
   } 
   \vspace{-5pt}
\label{fig:short_visualization_2}
\end{figure}

\begin{figure}[ht]
\begin{center}
\includegraphics[width=0.95\linewidth]{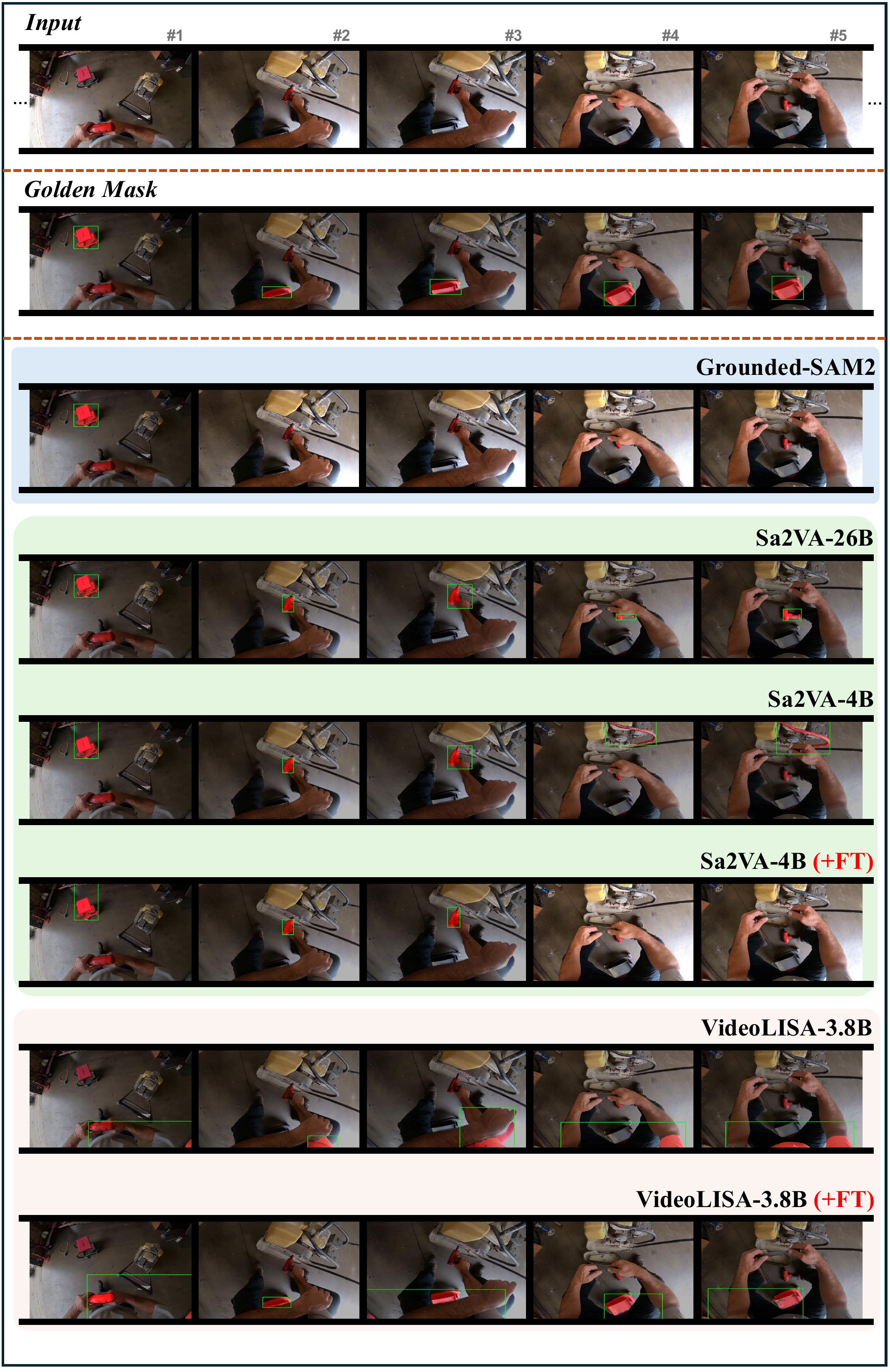}
\end{center}
\vspace{-15pt}
   \caption{Visualization of one example from \benchmark{}-Medium with sampled frames. The language query is ``the snap-on stool with a red cushion''.
   } 
   \vspace{-8pt}
\label{fig:mid_visualization}
\end{figure}

\begin{figure}[t]
\begin{center}
\includegraphics[width=0.95\linewidth]{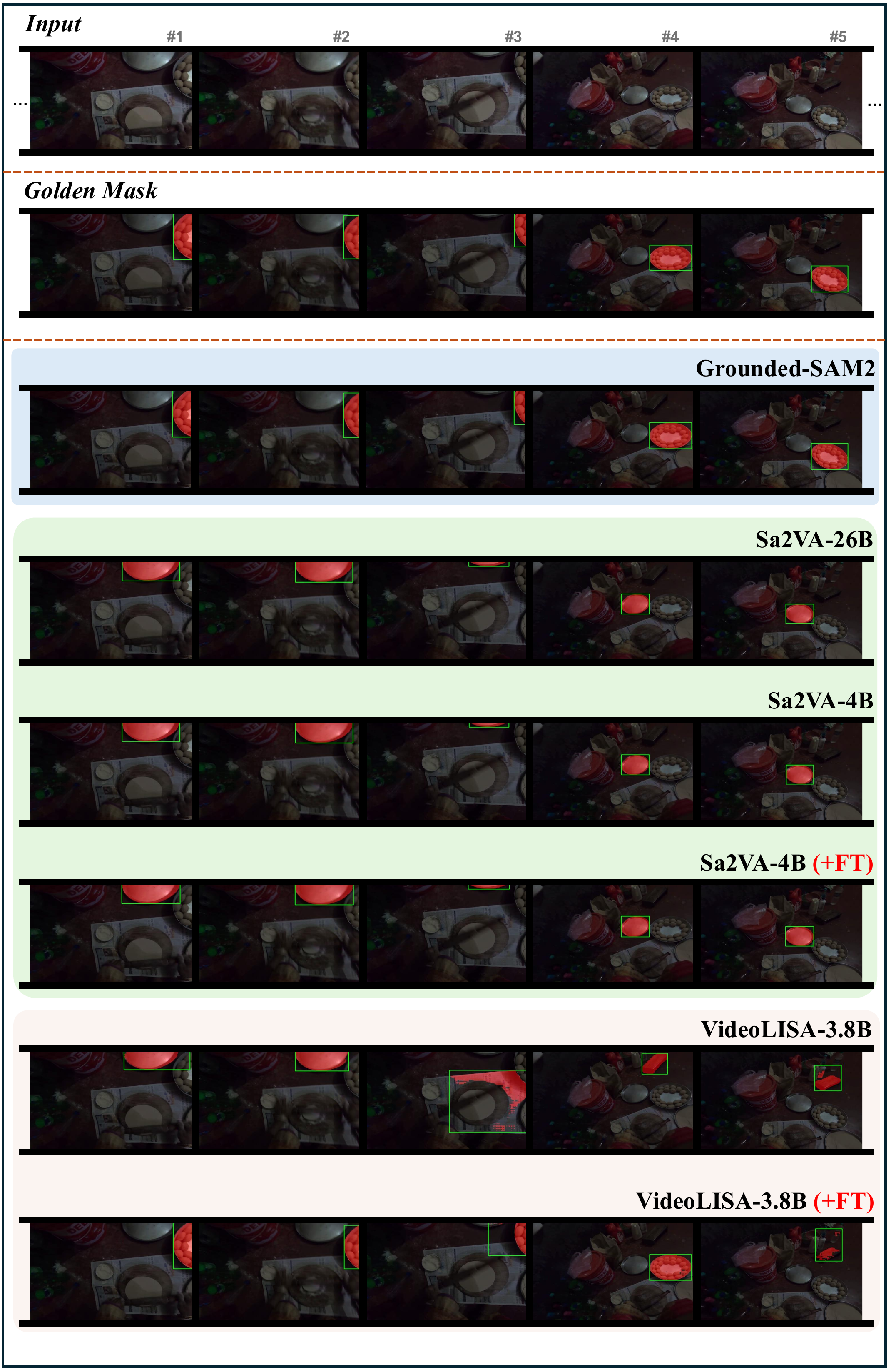}
\end{center}
\vspace{-10pt}
   \caption{Visualization of example from \benchmark{}-Medium. The language query is ``the circular silver-colored metal platter containing evenly arranged small, oval-shaped dough balls, placed on a table near a red container labeled "deepak" and surrounded by other kitchen items
''.
   } 
   \vspace{-5pt}
\label{fig:mid_visualization_2}
\end{figure}

\begin{figure}[t]
\begin{center}
\includegraphics[width=0.95\linewidth]{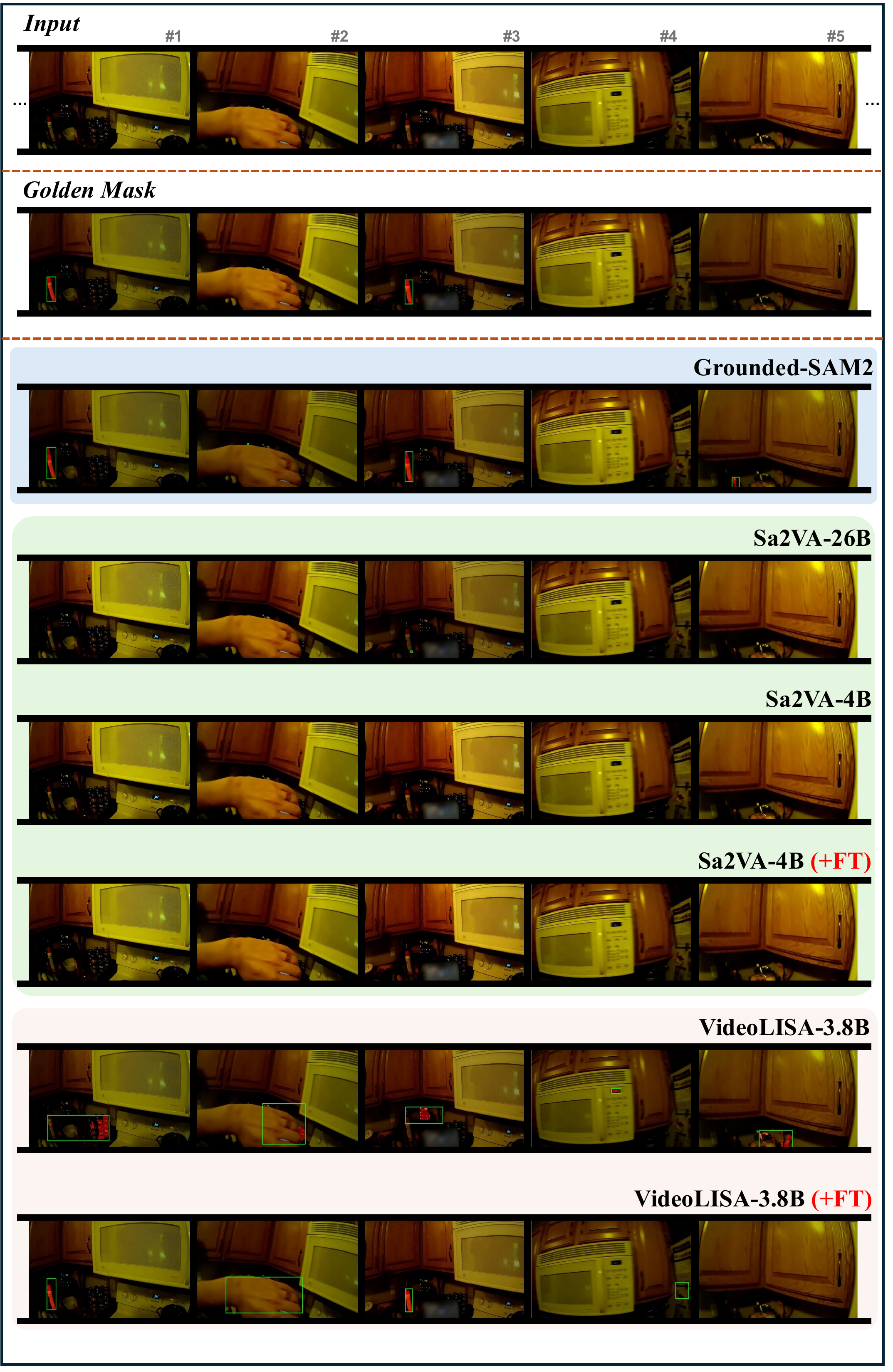}
\end{center}
\vspace{-10pt}
   \caption{Visualization of one example from \benchmark{}-Long. The language query is ``the tall, cylindrical white bottle with a red cap, located on the kitchen counter near the sink and surrounded by dishes and other kitchen items
''.
   } 
   \vspace{-5pt}
\label{fig:long_visualization_1}
\end{figure}

\section{More Visual Examples}
\label{sec:appendix_visual_examples}
We present more data examples from our proposed benchmark, along with the predictions from different grounding methods
in Figure~\ref{fig:short_visualization_2}, ~\ref{fig:mid_visualization}, ~\ref{fig:mid_visualization_2},~\ref{fig:long_visualization_1}.

\smallskip 
\noindent \textbf{Our fine-tuned models perform better than the pre-trained models}. After fine-tuning our proposed training dataset, the VideoLISA-3.8 (+FT) model can \textbf{avoid some grounding hallucinations} (~\#1 frames in Figure~\ref{fig:short_visualization_2}), \textbf{perform more precise grounding} (~\#2-~\#5 frames in Figure~\ref{fig:mid_visualization}, ~\#1-~\#4 frames in Figure~\ref{fig:mid_visualization_2}, and ~\#1/~\#3 frames in Figure~\ref{fig:long_visualization_1}).
Such performance improvements verify the effectiveness of our proposed training dataset \trainingdataset{}.

\smallskip 
\noindent \textbf{Query understanding ability matters}. The SAM2-based model has strong object-tracking ability. However, the capabilities of understanding the queries and knowing the correct object to ground are also important for spatiotemporal grounding tasks. Grounded-SAM2 has an inferior query understanding compared to VideoLLMs. As shown in Figure~\ref{fig:short_visualization_2}, it tracks the wrong object bed instead of the referred object pillow. 

All the above visual examples can show the difficulty of fine-grained spatiotemporal grounding on egocentric videos.

\end{document}